\documentclass[lettersize,journal]{IEEEtran}
\usepackage{amsmath,amsfonts}
\usepackage{algorithmic}
\usepackage{algorithm}
\usepackage{array}
\usepackage{textcomp}
\usepackage{stfloats}
\usepackage{url}
\usepackage{verbatim}
\usepackage{graphicx}
\usepackage{cite}
\usepackage{placeins}

\usepackage{bm} 
\usepackage{hyperref}

\usepackage{pgfplots}
\usepackage{tikz}
\usepackage{tikzscale}
\pgfplotsset{compat=1.3}
\pgfkeys{/pgf/number format/.cd,fixed}
\usetikzlibrary{plotmarks}
\usetikzlibrary{arrows.meta}
\usepgfplotslibrary{patchplots}
\newlength\fheight
\newlength\fheighta
\newlength\fheightb
\newlength\fheightc
\newlength\fwidth
\newlength\fwidtha
\newlength\fwidthb
\newlength\fwidthc

\usepackage{subcaption}

\newcommand{\skewcross}[1]{\lfloor #1 \rfloor^{\times}}
\makeatletter
\DeclareRobustCommand\frownotimes{\mathbin{\mathpalette\frown@otimes\relax}}
\newcommand{\frown@otimes}[2]{%
  \vbox{
    \ialign{##\cr
      \hidewidth$\m@th#1{}_\frown$\kern-\scriptspace\hidewidth\cr
      \noalign{\nointerlineskip\kern-1pt}
      $\m@th#1\otimes$\cr
    }%
  }%
}

\newcommand{\rev}[1]{\textcolor{black}{#1}}
\makeatother

\begin{document}

\title{Multi-IMU Sensor Fusion for Legged Robots}

\author{Shuo Yang,~\IEEEmembership{Student Member,~IEEE,} Zixin Zhang,~\IEEEmembership{Student Member,~IEEE,} John Z. Zhang,~\IEEEmembership{Student Member,~IEEE,} Ibrahima Sory Sow,~\IEEEmembership{Student Member,~IEEE,} Zachary Manchester,~\IEEEmembership{Member,~IEEE}
  \thanks{Shuo Yang, John Z. Zhang, Ibrahima Sory Sow, and Zachary Manchester are with the Robotics Institute and the Department of Mechanical Engineering, Carnegie Mellon University, Pittsburgh, PA, 15213, USA (e-mail: \href{mailto:shuoyang@andrew.cmu.edu}{shuoyang@andrew.cmu.edu}; \href{mailto:isow@andrew.cmu.edu}{isow@andrew.cmu.edu}; \href{mailto:ziyangz3@andrew.cmu.edu}{ziyangz3@andrew.cmu.edu}; \href{mailto:zmanches@andrew.cmu.edu}{zmanches@andrew.cmu.edu})}
  \thanks{Zixin Zhang is with the Department of Mechanical Engineering, Northwestern University, Evanston, IL, 60208, USA (e-mail: \href{mailto:zixinzhang2027@u.northwestern.edu}{zixinzhang2027@u.northwestern.edu} )}
}

\markboth{}%
{Shell \MakeLowercase{\textit{et al.}}: Multi-IMU Sensor Fusion for Legged Robots}


\maketitle
\begin{figure*}
  \centering
  \includegraphics[width=\linewidth]{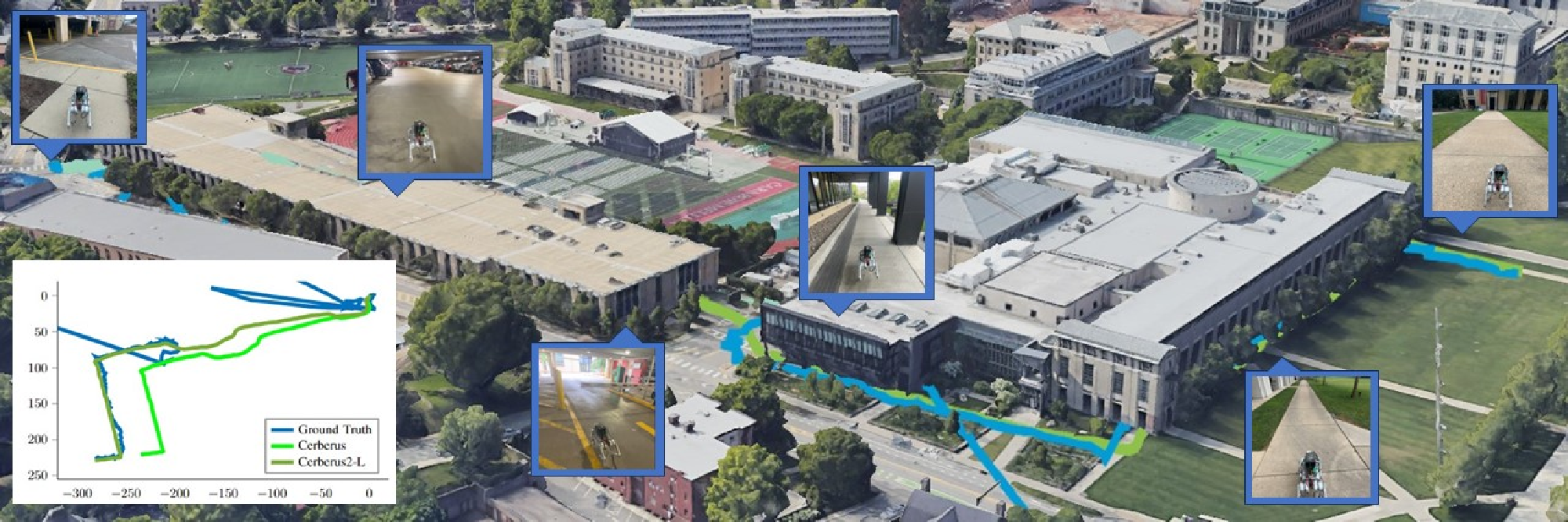}
  \caption{In an experiment run of over 500m, a Unitree Go1 robot travels through a parking garage over various terrain types. The proposed Cerberus 2.0 algorithm demonstrates great estimation robustness and accuracy.  Except for the Cerberus2 method and one baseline method (Cerberus), all other methods break along the way. The ground truth trajectory is also wrong inside the garage due to weak GPS signals. After the robot exited the garage, the GPS was restored. Cerberus2 has $<$ 0.5\% drift even after traveling for long distances with indoor/outdoor switching. The map can be viewed \href{https://www.google.com/maps/d/u/0/edit?mid=1bp65QUnbff6kFFlv7H4fXRPqdb5n6ps\&ll=40.44345689147336\%2C-79.94090941164713\&z=19}{online}.}
  \label{tro-fig:outdoor-garage}
\end{figure*}
\begin{abstract}
    \rev{This paper presents a state-estimation solution for legged robots that uses a set of low-cost, compact, and lightweight sensors to achieve low-drift pose and velocity estimation under challenging locomotion conditions. The key idea is to leverage multiple inertial measurement units on different links of the robot to correct a major error source in standard proprioceptive odometry. We fuse the inertial sensor information and joint encoder measurements in an extended Kalman filter, then combine the velocity estimate from this filter with camera data in a factor-graph-based sliding-window estimator to form a visual-inertial-leg odometry method.}
    We validate our state estimator through comprehensive theoretical analysis and
    hardware experiments performed using real-world robot data collected during a
    variety of challenging locomotion tasks. Our algorithm consistently achieves
    minimal position deviation, even in scenarios involving substantial ground
    impact, foot slippage, and sudden body rotations. A C++ implementation, along
    with a large-scale dataset, is available at
    \href{https://github.com/ShuoYangRobotics/Cerberus2.0}{https://github.com/ShuoYangRobotics/Cerberus2.0}. \end{abstract}

\begin{IEEEkeywords}
    Legged Robots, Sensor Fusion, Visual-Based Navigation, Field Robots
\end{IEEEkeywords}

\section{Introduction}

\rev{
    Long-term low-drift odometry has become an important research topic for legged
    robots in recent years. In many scenarios, legged robots must travel hundreds
    of meters autonomously \cite{tranzatto2022cerberus} and perform state
    estimation only using onboard sensors because external sensors that provide
    absolution position measurements, such as GNSS (Global Navigation Satellite
    Systems) and motion capture systems, are not always available. Therefore,
    global position estimates can only be obtained from the integration of linear
    velocity \cite{ma2016real}, which requires very precise velocity estimates.
    Moreover, since many legged robots have limited computational resources and
    limited payload capacity, visual-inertial-leg odometry (VILO)
    \cite{hartley2018hybrid, wisth2019robust}, where only cameras, inertial
    measurement units (IMUs), and leg-joint motor encoders are used for estimation,
    is an ideal choice. Compared to other sensor-fusion solutions,
    visual-inertial-leg odometry is small, lightweight, and low cost. }

%

\rev{Our goal is to develop a real-time odometry solution that generates high-frequency feedback for control purposes and also offers precise long-term position estimation. Although various visual-inertial odometry (VIO)~\cite{qin2018vins}, VILO \cite{wisth2022vilens}, and Lidar Odometry (LiO) \cite{zhang2015visual} solutions have been developed for general mobile robots and some legged robots, existing solutions often use a single IMU to provide inertial information. While legged robot state estimation solutions using multiple IMUs were explored before~\cite{elnecavexavier:hal-04382871}, it has not been integrated into VILO. In this work, we extend multi-IMU proprioceptive odometry \cite{yang2023multi} in several directions to improve estimator robustness and computational efficiency. We then integrate multi-IMU PO into a visual-inertial-leg odometry system that achieves low-drift position estimation performance across extensive indoor and outdoor experiments.
    To the best knowledge of the authors', our solution is the first open-source implementation of a multi-IMU visual-inertial-leg odometry method that achieves start-of-the-art estimation performance on legged robots. }

\rev{This paper extends two previous conference publications \cite{yang2022cerberus, yang2023multi} with the following new contributions:}
\begin{itemize}
    \item \rev{We improve the multi-IMU Proprioceptive Odometry (PO) proposed in~\cite{yang2023multi} by including foot orientations in the estimator state and derive analytical Jacobians of filter dynamics and measurement models.}
    \item \rev{We add a gravity direction measurement model to prevent body orientation from becoming unobservable \cite{bloesch2013state} when the robot stands still.}
    \item \rev{A detailed observability analysis is provided to confirm the filter's effectiveness.}
    \item \rev{We examine orientation estimation during fast rotation via the Cram\'{e}r-Rao Lower Bound~\cite{taylor1979cramer} to demonstrate the necessity of adding additional IMUs. }
    \item \rev{We provide an open-source implementation of our multi-IMU visual-inertial-leg odometry algorithm and all datasets used in the experimental section of the paper. }
\end{itemize}


\rev{Our contributions are as follows: First, we propose a compact and low-cost sensor suite design that uses additional IMUs on robot feet to account for rolling contact, foot slippage, and foot and ground deformation. In this way, a fundamentally incorrect assumption commonly used in legged odometry—that stance feet have zero velocities—can be dropped. Then we design the Multi-IMU PO that employs a Kalman filter to fuse multiple IMU and leg sensor data for pose and velocity estimation. The filter can also determine foot contact status through an outlier rejection mechanism. We derive analytical Jacobians for the filter dynamics and measurement models to preserve quaternion structures and achieve very efficient implementation. Further, we devise a sliding-window factor graph-based state estimator that combines information from the visual sensor, the inertial sensors, and the leg sensors, as well as the output of the Multi-IMU PO, to generate a refined pose and velocity estimation with higher accuracy and robustness. Moreover, we provide fully open-sourced code implementation with an easy-to-use, docker-based installation setup. A dataset containing sensor data collected during a total of 3 km and 1.5 hours of locomotion runs on a Unitree Go1 robot is made publicly available. The dataset covers various terrains, such as indoor flat ground, urban pedestrian walkways, parks, hiking trails, and so on.
}

This paper proceeds as follows: In Section \ref{tro-sec:related-work} we
provide an extensive literature review of relevant works. Section
\ref{tro-sec:background} provides mathematical notions and background knowledge
from legged robot state estimation. Section \ref{tro-sec:mipo-all} focuses on
the Multi-IMU Proprioceptive Odometry algorithm. \rev{Section
    \ref{tro-sec:cerberus2} presents Cerberus 2, which integrates the multi-IMU
    proprioceptive odometry in the factor graph framework.} Section
\ref{tro-sec:experiments} describes the implementation details and shows the
experiment results. Section \ref{tro-sec:discussion} discusses the limitations
and future research directions. Section \ref{tro-sec:conclusion} concludes the
paper.

\section{Related Work}\label{tro-sec:related-work}
In this section, we first provide brief surveys of the literature on
visual-inertial odometry and legged robot state estimation. Then we focus on
related work in visual-inertial-leg odometry.

\subsection{Visual-Inertial Odometry}

Using multiple sensors to estimate the physical state of a robot over long time
horizons is critical for autonomous robotics systems. Although \rev{GNSS can}
provide good position estimates, many robots need to operate in GNSS-denied
environments, such as under dense foliage, underground
\cite{tranzatto2022cerberus} or on other planetary bodies
\cite{cheng2005visual}. Visual odometry (VO) \cite{scaramuzza2011visual}, which
estimates robot pose using a monocular or a stereo pair of cameras, can provide
a solution in these settings. By matching features across image sequences,
feature locations constrain the possible motion of the camera so displacement
can be solved from multiple-view geometry \cite{hartley2003multiple}. The lidar
odometry \cite{zhang2014loam} leverages lidar sensors and shares a similar
principle as the VO. We focus on visual sensor-based methods because cameras
usually come with smaller form factors and lower costs. To improve robustness
and accuracy, VIO \cite{li2013high} uses information from both the camera and
the IMU as motion constraints. The VIO is usually realized by Kalman filtering
\cite{kalman1960new} or a factor graph \cite{dellaert2012factor}, both of which
can be viewed as solving optimization problems involving motion constraints as
cost functions \cite{humpherys2012fresh}. The factor graph provides a flexible
representation to enable VIO to keep track of a large number of visual features
\cite{kaess2008isam}, exploit problem sparsity \cite{dellaert2006square}, and
deal with sensors with different frequencies \cite{forster2015imu}. Moreover,
the factor graph can incorporate loop closure \cite{thrun2005probabilistic,
    engel2014lsd, mur2015orb} easily where the position estimation drift can be
greatly reduced if the robot visits a previous location again. Dealing with
loop closures at different time scales is one of the central topics of
Simultaneous Localization And Mapping (SLAM) In this work, we do not consider
loop closure because our goal is to study cumulative odometer position drift.

To achieve low drift in VO or VIO, several VO or VIO algorithms
\cite{li2013high, sun2018robust, qin2018vins} consider IMU biases, sensor
delays \cite{li2014online}, and extrinsic parameter errors \cite{qin2018online}
within the odometer algorithms. It has been shown that properly addressing
these error sources can improve odometry performance. As an important
performance metric, position drift percentage, dividing final drift distance by
total travel distance, is often used \cite{shen2013vision}. With good
engineering efforts and having error sources properly addressed, the position
drift of a VIO estimator can be as low as 0.29\% on drones \cite{qin2018vins}
or 0.32\% on cars \cite{huai2022robocentric}.
\subsection{Legged Robot State Estimation}
Research on onboard real-time state (including mainly pose and velocity)
estimation for legged robots became popular in the recent decade
\cite{bloesch2013state,bledt2018cheetah, rotella2014state, kim2021legged,
    wisth2019robust, hartley2018legged, yang2022cerberus}. With the increasing
commercial availability of low-cost quadrupedal \cite{a1, spotmini} and
humanoid robots, there is a strong need for cost-effective and reliable sensing
solutions for such robots with limited computation resources.

Leg Odometry (LO) is the earliest state estimation method for humanoid and
hexapod robots \cite{kajita2014introduction, lin2005leg}. The robot pose can be
calculated using the leg kinematics from feet that are in stable contact with
the ground. Repeated dead reckoning using the stance foot allows the position
trajectory to be estimated \cite{roston1991dead}. However, LO itself is prone
to errors such as foot slippage and joint encoder noise.
\cite{reinstein2011dead, skaff2010context} combines IMU and LO using Extended
Kalman Filters (EKF) to improve accuracy and robustness. An observability
analysis presented in \cite{bloesch2013state} shows that, on a 12 DOF quadruped
robot, \rev{the body pitch and roll angles}, velocity, and IMU biases can be
recovered using one body IMU, joint encoders, and foot contact sensors.
\rev{Still, the position and yaw angle are not observable.} This EKF
formulation is also applicable to humanoid robots \cite{rotella2014state}. A
similar linear KF formulation is proposed in \cite{bledt2018cheetah}. Invariant
EKF is proposed in \cite{hartley2020contact} to improve convergence of
orientation estimation. Contact sensing \cite{camurri2017probabilistic,
    bloesch2013slip} is critical to ensure that LO is only used while feet are in
contact. Some algorithms estimate contacts using kinematic information
\cite{hwangbo2016probabilistic}, eliminating the dependency on foot contact
sensors. Since all sensors are proprioceptive, in other words, measuring only
the internal state of the robot, these methods all belong to Proprioceptive
Odometry (PO).

In PO methods, velocity estimations are typically good enough for stable
closed-loop control, but position drifts are often as high as 10\%-15\%
\cite{bloesch2013state, hartley2020contact, kim2021legged,
    wisth2020preintegrated}. The reason is that, in addition to foot slippages, a
legged robot also often experiences link and ground deformations, rolling
ground contacts \cite{yang2022online}, and excessive impacts with the ground,
all of which lead to either large sensor noise or incorrect or biased velocity
estimation if sensor measurement models do not capture the actual contact
behavior. It has been shown that properly addressing these issues in PO can
improve the accuracy of position estimation \cite{yang2022online}.
Nevertheless, PO is widely used because not all legged robot control
applications require high-accuracy position estimation.

Instead of polishing details within the PO, another way to improve estimation
accuracy for legged robots is to introduce cameras and lidars
\cite{kuindersma2016optimization, ma2016real}. This combines PO and VIO and
extends them to Visual-inertial-leg Odometry (VILO) \cite{hartley2018hybrid,
    wisth2019robust, kim2021legged, yang2022cerberus}, which can be implemented in
different ways. Prior work has shown that combining three to four types of
sensors commonly available on all legged robots to do state estimation can
significantly improve result accuracy over methods that just use subsets of
sensors \cite{wisth2022vilens, yang2022cerberus}. In the next section, we
discuss several VILO realizations that exist in the literature.

\subsection{Visual-inertial-leg Odometry}
A key concept commonly used to classify VIO is coupling \cite{shen2013vision,
    li2014visual, scaramuzza2019visual, cioffi2020tightly}, which is also crucial
in the VILO discussion. A \textit{loosely-coupled} odometer first generates
pose estimations from two or more sub-modules individually, then aligns them
using a certain optimization routine~\cite{wisth2019robust}. In this approach,
knowledge of how to correct error sources in one submodule is not shared with
others, resulting in potentially suboptimal estimation results. While the
\textit{tightly-coupled} approach considers all sensor information within a
single optimization program \cite{wisth2022vilens}, which could achieve higher
precision by reaching joint optimality.

The loosely-coupled approach can be done by either adding VIO output velocity
into Kalman filter-based PO as a measurement model
\cite{kuindersma2016optimization} or adding PO velocity output into the VIO as
additional factors \cite{hartley2018legged}. A loosely-coupled KF solution
using a tactical-grade IMU and high-quality cameras with an FPGA-based sensor
synchronization mechanism to achieve less than 1\% position estimation drift on
the Boston Dynamics LS3 robot was built in \cite{ma2016real}. The factor
graph-based VILO presented in \cite{wisth2019robust} has a KF
PO~\cite{bloesch2017two} and a factor graph VIO as submodules, the pose
displacement estimated by the PO is added into the VIO as additional motion
constraints.

The tightly-coupled VILO uses a factor graph and its underlying optimization
framework to process information from all sensors together
\cite{wisth2022vilens, yang2022cerberus}. To deal with the fact that IMU and
leg sensors usually have much higher frequencies than cameras, contact
preintegration is developed in \cite{hartley2018hybrid} following the IMU
preintegration \cite{forster2015imu}. Like IMU biases can be estimated during
IMU preintegration, \cite{wisth2020preintegrated} describes the LO velocity
bias and models it as a linear term that can be corrected in the contact
preintegration. However, this bias model does not explain the source of the
bias or its physical meaning. It is further improved by adding a lidar sensor
to achieve 0.2\%-0.4\% drift performance \cite{wisth2022vilens}, though their
VILO implementation and datasets are not publicly available.

\subsection{\rev{Multi-IMU State Estimation}}
\rev{Lastly, we note that deploying a suite of IMUs, rather than just one, to enhance estimation performance is common in several domains, such as pedestrian navigation systems \cite{bancroft2011data}, augmented and virtual-reality applications \cite{jadid2019utilizing}, and general VIO \cite{eckenhoff2021mimc}
    In legged robot state estimation, a PO solution is proposed in
    \cite{xinjilefu2016distributed} using a network of IMUs installed on different
    parts of a humanoid robot for better joint-velocity sensing. A similar approach
    uses redundant accelerometers to improve joint information
    estimation~\cite{leboutet2020second}. A multi-IMU KF is developed
    \cite{elnecavexavier:hal-04382871} for an exoskeleton. It is also shown that
    multiple IMUs can help quantify link flexibilities~\cite{vigne2019state}.
    However, multiple IMUs have not previously been combined with VILO. }




\section{Background}\label{tro-sec:background}
\rev{We now introduce notation and review mathematical background material that is used in subsequent sections.}
We begin with some general notation: Lowercase letters are used for scalars and coordinate frame abbreviations. Boldface lowercase letters are vectors. Uppercase letters are matrices and vector sets. The operation $[a;b;c]$ vertically concatenates elements $a$, $b$ and $c$. 
The operator $\skewcross{\bm{v}}$ converts a vector $\bm{v} = [v_1;v_2;v_3]\in \mathbb{R}^3$ into the skew-symmetric ``cross-product matrix,''
\begin{equation}
    \skewcross{\bm{v}} =
    \begin{bmatrix}
        0    & -v_3 & v_2  \\
        v_3  & 0    & -v_1 \\
        -v_2 & v_1  & 0
    \end{bmatrix} ,
\end{equation}
such that $\bm{v} \times \bm{x} = \skewcross{\bm{v}} \bm{x}$. $\dot{\bm{a}}$ is the time derivative of $\bm{a}$. The Mahalanobis norm \cite{chandra1936generalised} $\|\bm{a}\|_P$ calculates a scalar value equals to $\sqrt{\bm{a}^TP^{-1}\bm{a}}$.
Lastly, $\hat{\bm{a}}$ indicates the estimated mean value of a random variable follows the Gaussian distribution $\mathcal{N}(\bm{a},\Sigma)$.
\subsection{Coordinate Frames}
\begin{figure}
    \centering
    \includegraphics[width=0.7\linewidth]{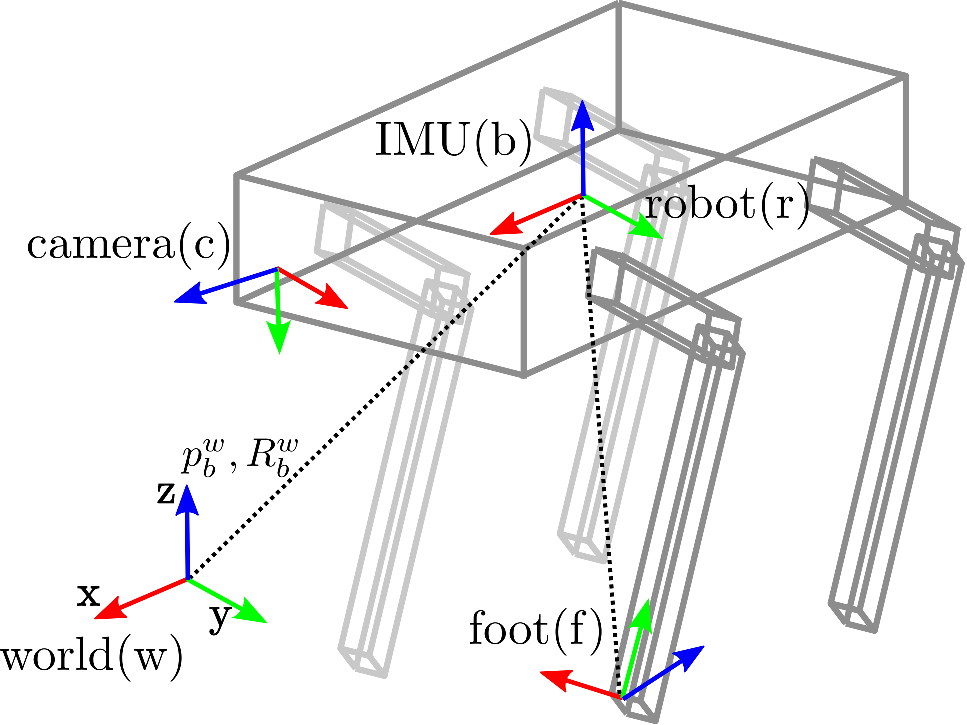}
    \caption{Coordinate frame definitions used for a typical quadrupedal robot.}
    \label{fig:frame}
\end{figure}
Important coordinate frames are shown in Fig. \ref{fig:frame}. For simplicity, we assume that the IMU frame and the robot's body frame coincide. \rev{We use $\bm{p} \in \mathbb{R}^3$ to denote the translation vector and $R \in SO(3)$ to denote the rotation matrix from the robot's body frame to the world frame~\cite{lynch2017modern}.} The matrix $R$ can also be viewed as a representation of robot orientation.
Where necessary, we use superscripts and subscripts to explicitly indicate the frames associated with rotation matrices and vectors, so $R^{a}_{b}\cdot \bm{p}_b$ means the matrix transforms a vector $\bm{p}_b$ represented in coordinate frame $b$ into coordinate frame $a$ \cite{murray2017mathematical}. For brevity, if frame $b$ is time-varying, we write $R^w_k$ instead of $R^{w}_{b(k)}$ to indicate the rotation matrix is also time dependent. Similarly, $\bm{p}_{k}$ defines a time-varying vector, it can also be viewed as the origin vector of frame $b_k$ in the world.

\subsection{Quaternions \& Rotations}
A rotation matrix can be parametrized alternatively as a unit quaternion. We
follow the quaternion convention defined in \cite{jackson2021planning}. A
quaternion $\bm{q} = [q_s; \bm{q}_v]$ has a scalar part $q_s$ and a vector part
$\bm{q}_v = [q_x; q_y; q_z] \in \mathbb{R}^3$. A unit quaternion is the one
that has $\|\bm{q}\| = 1$. We define the following two matrices
\begin{align}
    \mathcal{L}(\bm{q}) & =
    \begin{bmatrix}
        q_s      & -\bm{q}_v^{\top}          \\
        \bm{q}_v & q_sI+\skewcross{\bm{q}_v}
    \end{bmatrix}
    \ \text{and}  \label{tro-eqn:left-quat-map} \\
    \mathcal{R}(\bm{q}) & =
    \begin{bmatrix}
        q_s      & -\bm{q}_v^{\top}          \\
        \bm{q}_v & q_sI-\skewcross{\bm{q}_v}
    \end{bmatrix},\label{tro-eqn:right-quat-map}
\end{align}
such that the product of two quaternions can be written as
\begin{equation}
    \bm{q}_1 \otimes \bm{q}_2 = \mathcal{L}(\bm{q}_1)\bm{q}_2 = \mathcal{R}(\bm{q}_2)\bm{q}_1.
\end{equation}
It can also be shown that the inverse of a unit quaternion $\bm{q}$ is $\bm{q}^{-1} = [q_s;-\bm{q}_v]$ and $\bm{q}\otimes\bm{q}^{-1}=\bm{q}_I = [1;\bm{0}]$, the identity quaternion.
We also introduce a matrix $B = \begin{bmatrix}
        0 \\I_{3\times3}
    \end{bmatrix}$ that converts a vector in $\mathbb{R}^3$ to a quaternion with zero scalar part. We denote function $R(\bm{q}) \in SO(3)$ as a mapping from a unit quaternion $\bm{q}$ to a rotation matrix, which is in the form of
\begin{equation}
    R(\bm{q}) = B^{\top}\mathcal{L}(\bm{q})\mathcal{R}(\bm{q})^{\top}B.
\end{equation}

\subsection{\rev{Small Rotation Representations}}\label{tro-sec:small-angle}
Small rotation approximations play an important role in orientation estimation. \rev{Assuming that the true body orientation of a robot is $\bm{q}$ and our estimate of it is $\hat{\bm{q}}$, we define the orientation error as $\delta \bm{q} = \hat{\bm{q}}^{-1}\otimes\bm{q}$. To facilitate the development of estimation algorithms, standard practice is to convert the orientation error into a three-parameter representation using the logarithmic map~\cite{barfoot2024state} or the inverse Cayley map~\cite{jackson2021planning}. Similarly, the exponential map or the Cayley map can convert a three-parameter vector to a rotation. For a unit quaternion $\bm{q}$ and a three-parameter representation vector $\bm{\theta} = \phi\mathbf{u}\in \mathbb{R}^3$, where $\phi$ is a scalar and $\mathbf{u}$ is a unit vector, the exponential and logarithmic maps are defined as~\cite{altmann2005rotations}}
\begin{align}
    \text{Exp}(\bm{\theta}) & = \begin{bmatrix}
                                    \cos(\frac{\phi}{2}) \\
                                    \sin(\frac{\phi}{2})\mathbf{u}
                                \end{bmatrix}, \text{and}                                             \\
    \text{Log}(\bm{q})      & = \phi\mathbf{u} = 2\arctan(\|\bm{q}_v\|, q_s) \cdot \bm{q}_v/\|\bm{q}_v\|.
\end{align}
\rev{In this work, we use approximations of the exponential and logarithm maps common in spacecraft attitude estimation~\cite{lefferts1982kalman} and visual-inertial odometry~\cite{qin2018vins}:}
\begin{align}
    \text{Exp}(\bm{\theta}) & \approx \begin{bmatrix}
                                          1 \\
                                          \frac{\phi}{2} \mathbf{u}
                                      \end{bmatrix}, \label{tro-eqn:quat-exp-map}               \\
    \text{Log}(\bm{q})      & = \phi\mathbf{u} \approx 2 \bm{q}_v. \label{tro-eqn:quat-log-map}
\end{align}
\rev{We also make use of a similar first-order approximation of the matrix exponential~\cite{hall2015lie}: }
\begin{align}
    \text{Exp}(\bm{\theta}) \approx (I+\skewcross{\bm{\theta}}) \label{tro-eqn:rot-exp-map}.
\end{align}


\subsection{Forward Kinematics \& Leg Odometry Velocity}
In this section, we review the forward kinematics of a robot's leg and describe
how to infer the robot's body velocity from joint-angle information. For the
$j$th leg of a legged robot, we define $\bm{\alpha}$ as a vector containing all
joint angles. The forward kinematics function is denoted $ \bm{p}^b_{f} =
    \bm{g}(\bm{\alpha}) \in \mathbb{R}^3 $, whose output is the foot position in
the robot body frame. The derivative of this equation leads to the Jacobian
matrix $J(\bm{\alpha})$ that maps $\dot{\bm{\alpha}}$ into the foot's linear
velocity in the body frame:
\begin{equation}
    \bm{v}^b_{f} = \dot{\bm{p}}^b_{f} = J(\bm{\alpha})\dot{\bm{\alpha}}.
\end{equation}

Let $\bm{p}^w_f$ denote the foot position in the world frame, then
\begin{equation}\label{tro-eqn::foot_position}
    \bm{p}^w_f = \bm{p} + R(\bm{q})\bm{p}^b_{f} = \bm{p} + R(\bm{q})\bm{g}(\bm{\alpha}),
\end{equation}
where $\bm{p}$ and $\bm{q}$ are the robot body pose in the world frame. Denote by $\dot{\bm{p}}^w_f$, the time derivative of $\bm{p}^w_f$, $\bm{v}^w_f$. By differentiating \eqref{tro-eqn::foot_position}, we have:
\begin{equation}\label{tro-eqn::foot_position_derive}
    \bm{v}^w_f = \dot{\bm{p}}^w_f = \dot{\bm{p}} + R(\bm{q})\frac{d}{dt}\bm{g}(\bm{\alpha}) + \frac{d}{dt}R(\bm{q})\bm{g}(\bm{\alpha}).
\end{equation}
It is shown in \cite{murray2017mathematical} that $\frac{d}{dt}R(\bm{q}) = R(\bm{q})\skewcross{\bm{\omega}_b}$, where $\bm{\omega}_b$ is the robot body angular velocity. We also write $\bm{v} = \bm{\dot{p}}$, then
from \eqref{tro-eqn::foot_position_derive} we derive an expression for the body
velocity in the world frame:
\begin{equation}
    \label{tro-eqn:lo}
    \bm{v} = -R(\bm{q})[ J(\bm{\alpha})\dot{\bm{\alpha}} + \skewcross{\bm{\omega}_b}\bm{g}(\bm{\alpha})] + \bm{v}^w_f.
\end{equation}

This velocity is called the LO velocity because its integration is the body
displacement \cite{lin2005leg}.

A fundamental assumption called the ``non-slip'' or ``zero-velocity''
assumption commonly used in the legged robot state estimator is treating
$\bm{v}^w_f = 0$ for a foot that contacts the ground. As we are going to show,
this assumption can lead to large position estimation drift because the foot
velocity during contact may not be zero.

\subsection{Extended Kalman Filter}\label{tro-sec:ekf}
The Extended Kalman Filter (EKF) is widely used in robot state estimation. The
basic building blocks are a dynamics or ``process'' model:
\begin{align}
    \bm{x}_{k+1}=\bm{f}(\bm{x}_k) + \bm{n}_k
\end{align}
and a measurement model:
\begin{align}
    \bm{z}_k=\bm{h}(\bm{x}_k) + \bm{w}_k,
\end{align}
where $\bm{n}_k$ and $\bm{w}_k$ are zero-mean additive noise terms drawn from Gaussian distributions with covariances $\Sigma_n$ and $\Sigma_w$, respectively. We also make use of the model Jacobians $F_k = \partial \bm{f} / \partial \bm{x}_k$ and $H_k = \partial \bm{h} / \partial \bm{x}_k$ \cite{maybeck1982stochastic}.

Given an estimated state $\hat{\bm{x}}_{k}$ and its associated covariance
$P_{k}$ at time $k$ and a sensor measurement vector $\bar{\bm{z}}_k$, the EKF
first ``predicts'' the mean of the state distribution at time $k+1$ using the
process model:
\begin{align}
    \hat{\bm{x}}_{k+1|k} & = \bm{f}(\hat{\bm{x}}_{k}),          \\
    P_{k+1|k}            & = F_{k+1} P_{k}F_{k+1}^T + \Sigma_n,
\end{align}
where the Jacobin $F_{k+1}$ is evaluated at $\hat{\bm{x}}_{k+1|k}$.
The filter then performs a ``measurement update'', in which a measurement residual $\bm{y}_{k+1} = \bar{\bm{z}}_k - h(\hat{\bm{x}}_{k+1|k})$ is formed and the Kalman gain \cite{kalman1960new} $K_{k+1}$ is calculated as:
\begin{align}
    S_{k+1} & = H_{k+1}P_{k+1|k}H_{k+1}^T + \Sigma_w \label{tro-eqn:ekf-S}, \\
    K_{k+1} & = P_{k+1|k}H_{k+1}^TS_{k+1}^{-1}.
\end{align}
Then the updated estimated state distribution is calculated as:
\begin{align}
    \hat{\bm{x}}_{k+1} & = \hat{\bm{x}}_{k+1|k} + K_{k+1}\bm{y}_{k+1},  \label{tro-eqn:ekf-update-x} \\
    P_{k+1}            & = (I-K_{k+1}H_{k+1})P_{k+1|k}.  \label{tro-eqn:ekf-update-P}
\end{align}
These two quantities describe the estimated Gaussian distribution of the state at time $k+1$. We refer the reader to standard references on state estimation for more details \cite{maybeck1982stochastic}.

\rev{If the orientation state is parameterized by unit quaternions, special care must be taken to preserve the quaternion group structure during the EKF update. First, the Jacobians should be computed with respect to the small rotation introduced in Section~\ref{tro-sec:small-angle}~\cite{madyastha2011extended, yang2022online}. Second, the state update equation \eqref{tro-eqn:ekf-update-x} will use the quaternion exponential map instead of the normal addition. }


A common outlier rejection technique used in Kalman filtering is to compare the
Mahalanobis norm between estimated state and measurement
\cite{vaganay1996outlier}. We will leverage this technique to improve our
filter's robustness, as will be discussed in Section
\ref{tro-sec:mipo-outlier}.

\subsection{Standard Single-IMU Proprioceptive Odometry}\label{tro-sec:sipo}

Let the robot's state be ${\bm{x}} = [{\bm{p}}; {\bm{v}}; {\bm{q}};
    {\bm{s}}_{1},\dots,{\bm{s}}_{j},\dots,{\bm{s}}_{E}]$, where ${\bm{p}}\in
    \mathbb{R}^3$ is the robot position in the world frame, ${\bm{q}}$ is the
robot's orientation quaternion, and ${\bm{v}} \in \mathbb{R}^3$ is the linear
velocity of the robot's body represented in the world frame. For
$j\in\{1,\dots, E\}$ where $E$ is the total number of legs of the robot,
${\bm{s}}_{j}$ is the foot position of leg $j$ represented in the world frame.
For clarity, we will only discuss the case when $E=1$ in this section and drop
the symbol $j$ from subsequent equations. The robot's sensors generate several
measurements including IMU acceleration $\bm{a}_{b}$, IMU angular velocity
$\bm{\omega}_{b}$, joint angle $\bm{\alpha}$, joint-angle velocity
$\dot{\bm{\alpha}}$, and $c$ which is a binary contact flag with $c = 1$
indicating foot contacts with the ground. To keep the background introduction
brief, we defer the discussion of IMU bias to later sections and assume IMU
readings in this section to be bias-free.

The standard PO uses the EKF to estimate the state from a single IMU and the LO
velocity \cite{bloesch2013state, bledt2018cheetah}.
The process model of the standard PO is based on IMU kinematics. A
discrete-time prediction model using Euler integration is presented in
\cite{bledt2018cheetah} as:
\begin{align}
    \hat{\bm{x}}_{k+1|k} =
    \begin{bmatrix}
        \hat{\bm{p}}_{k+1|k} \\
        \hat{\bm{v}}_{k+1|k} \\
        \hat{\bm{q}}_{k+1|k} \\
        \hat{\bm{s}}_{k+1|k}
    \end{bmatrix}=
    \begin{bmatrix}
        \hat{\bm{p}}_{k} + \Delta t\hat{\bm{v}}_{k}                       \\
        \hat{\bm{v}}_{k} + \Delta t(R(\hat{\bm{q}}_k)\bm{a}_b - \bm{g}^w) \\
        \mathcal{L}(\hat{\bm{q}}_{k})\text{Exp}(\bm{\omega}_b\Delta t)    \\
        \hat{\bm{s}}_{k}
    \end{bmatrix}\label{tro-eqn:baseline-process},
\end{align}
where
$\Delta t$ is the time interval between $k$ and $k+1$. $\bm{g}^w$ is the gravity vector.

A common technique used in the literature is contact-based covariance
scaling~\cite{bloesch2013state, bledt2018cheetah}. Since we cannot update the
foot position in the process model \cite{bloesch2013state} during foot swing,
for the term corresponding to $\hat{\bm{s}}$ in $\bm{n}_k$, we set its
covariance $\sigma_{s}$ to a large value if $c = 0$, and a small value
otherwise:
\begin{align}
    \sigma_{s} & =  c \sigma_c +  (1-c)\sigma_{n} \label{tro-eqn:covariance-enlarge}.
\end{align}
$\sigma_c$ and $\sigma_{n} >> \sigma_c$ are all tunable hyperparameters. This technique works well in practice, but it will increase the condition number of the covariance matrix, thus reducing the stability of the filter~\cite{maybeck1982stochastic}.

We formulate the EKF measurement model following \cite{bledt2018cheetah}. From
sensor measurements, a vector $\bar{\bm{z}}_k$ is obtained as:
\begin{align}
    \bar{\bm{z}}_k = \begin{bmatrix}
                         \bm{g}(\bm{\alpha}) \\-J(\bm{\alpha})\dot{\bm{\alpha}}+\skewcross{\bm{\omega}_{b}}\bm{g}(\bm{\alpha})
                     \end{bmatrix}\label{tro-eqn:baseline-measurement}
\end{align}
The measurement function $\bm{h}(\hat{\bm{x}}_k)$ is defined as:
\begin{align}
    \bm{h}(\hat{\bm{x}}_k) =  \begin{bmatrix}
                                  R(\hat{\bm{q}}_k)^T(\hat{\bm{s}}_k-\hat{\bm{p}}_k) \\
                                  R(\hat{\bm{q}}_k)^T\hat{\bm{v}}_k
                              \end{bmatrix}\label{tro-eqn:baseline-residual}
\end{align}
The first term of the residual $\bar{\bm{z}}_k -\bm{h}(\hat{\bm{x}}_k)$ indicates that the estimated body position and foot position must differ by a distance equal to the leg forward kinematics position. The second term ensures that the estimated robot body velocity matches the LO velocity \eqref{tro-eqn:lo} with the assumption that $\bm{v}^w_f=0$, which we refer to as a ``zero-velocity'' observation model.
We can utilize residuals from all legs in the EKF by stacking them into a single vector. However, measurement residuals are only meaning for non-slipping contact feet. Therefore, the noise covariance $\Sigma_w$ is adjusted based on the contact flag $c$ as in equation \eqref{tro-eqn:covariance-enlarge} \cite{bloesch2013state}.

The robot state can include additional terms such as IMU biases
\cite{savely1972apollo}, sensor delay time \cite{li2014online}, and leg
kinematic parameters \cite{yang2022online}. Estimating them along with robot
pose and velocity can improve the overall estimation accuracy
\cite{maybeck1982stochastic}.

\subsection{Visual-Inertial-Leg Odometry}\label{tro-sec:vilo}
Modern VILO methods mainly adopt a nonlinear factor graph
\cite{dellaert2017factor, thrun2005probabilistic} optimization
approach~\cite{hartley2018hybrid,wisth2019robust, yang2022cerberus}. It has a
close connection to the Kalman filter.

From an optimization perspective, the Kalman filter presented in Section
\ref{tro-sec:ekf} is equivalent to solving a nonlinear least square problem:
\begin{align}
    \min_{\bm{x}_{k+1},\bm{x}_k} & \|\bm{x}_k-\hat{\bm{x}}_k\|^2_{P_k} + \nonumber                                                                                       \\
                                 & \|\bm{x}_{k+1}-\bm{f}(\bm{x}_k)\|^2_{\Sigma_n} + \|\bar{\bm{y}}_{k+1}-\bm{h}(\bm{x}_{k+1})\|^2_{\Sigma_w} \label{tro-eqn:kf-opt-form}
\end{align}
followed by a marginalization step \cite{humpherys2012fresh}. In the above optimization problem, each term is in the least square form, representing a ``constraint residual'' that indicates the best possible values of decision variables. We can generalize Problem \eqref{tro-eqn:kf-opt-form} to include a history of past states and related constraint residuals and lead to the Kalman smoother\cite{humpherys2012fresh}. Moreover, if we discard the Markov assumption \cite{maybeck1982stochastic}, terms in the least square problem can involve not only states from adjacent time steps. For example, a landmark may be seen by a camera sensor from multiple locations a robot visited, so a camera measurement is related to all the robot states at these visit times. A factor graph \cite{dellaert2012factor} captures the relationships among decision variables and their involvements in residual terms graphically.

We introduce the general form of a factor graph-based VILO framework
\cite{wisth2020preintegrated, hartley2018hybrid, kim2022step}. More
specifically, a VILO keeps track of the estimation of a list of past $N$ robot
states $\hat{\bm{x}}_k$ and $K$ camera feature landmark locations
$\hat{\lambda}_l$ as $\mathcal{X} = \{\hat{\bm{x}}_0, \hat{\bm{x}}_1, \dots
    \hat{\bm{x}}_N, \hat{\lambda}_0, \hat{\lambda}_1, \dots \hat{\lambda}_K\}$. The
robot state is $\hat{\bm{x}}_k = [\hat{\bm{p}}_k; \hat{\bm{q}}_k;
    \hat{\bm{v}}_k; \hat{\bm{b}}_{ak}; \hat{\bm{b}}_{\omega k}]$, where
$\hat{\bm{p}}_k \in \mathbb{R}^3$ is the robot position in the world frame,
$\hat{\bm{q}}_k$ is the robot's orientation quaternion and $\hat{\bm{v}}_k \in
    \mathbb{R}^3$ is the linear velocity of the robot's body represented in the
world frame. $\hat{\bm{b}}_{ak} \in \mathbb{R}^3$ and $\hat{\bm{b}}_{\omega k}
    \in \mathbb{R}^3$ are IMU accelerometer bias and gyroscope bias. A new state
$\hat{\bm{x}}_k$ is created each time $t_k$ when a new camera image arrives.
Also, sensors on the robot generate measurements
$Z_t=\{\bm{a}_{b},\bm{\omega}_{b},\bm{\alpha}_{j},\dot{\bm{\alpha}}_{j},c\}$
and $\Lambda_{t}$ periodically, where measurements in $Z_t$ are introduced in
Section \ref{tro-sec:sipo}
and $\Lambda_{t}$ is a set of feature coordinates in the camera images that
have known associations with feature locations in $\mathcal{X}$. We denote
$\mathcal{Z}$ as all measurements between state $\hat{\bm{x}}_0$ and
$\hat{\bm{x}}_N$. We also denote subsets $\mathcal{X}_{sub} \subset
    \mathcal{X}$ and $\mathcal{Z}_{sub} \subset \mathcal{Z}$. VILO constructs a
nonlinear least-squares problem to find $\mathcal{X}$ as the solution of

\begin{equation}\label{tro-eqn:map}
    \min_{\mathcal{X}} \bigg\{ \sum_i \bigg\| \bm{r}_i(\mathcal{X}_{sub}, \mathcal{Z}_{sub})\bigg\|^2_{P_i} \bigg\},
\end{equation}
where each term $\bm{r}_i(\mathcal{X}_{sub}, \mathcal{Z}_{sub})$ is a general constraint residual formulation that involves a certain number of states and sensor measurements. $P_i$ is a weighting matrix that encodes the relative uncertainty in each $\bm{r}_i$, whose value depends on the constraint type. For example, all terms in \eqref{tro-eqn:kf-opt-form} can be seen as in such residual formulations. Ideally, the gradient of the total residual should be \textbf{0} at optimal solution $\mathcal{X}^*$.  Problem \eqref{tro-eqn:map} can be solved by nonlinear optimization methods \cite{dellaert2017factor}. The core technical challenge of constructing a VILO is to design residual functions leveraging all available sensor data.
\rev{Additionally, a VILO estimator usually has other mechanisms to ensure real-time computation, such as camera feature tracking management, preintegration, and sliding window marginalization. See \cite{qin2018vins, wisth2019robust} for more details. }

\section{Multi-IMU Kalman Filtering}\label{tro-sec:mipo-all}

The standard single-IMU PO method presented in Section \ref{tro-sec:sipo} has
two fundamental limitations: First, bias estimation, which usually must be
included if IMU hardware quality is not good, would drift when the robot stands
still with colinear contacts \cite{bloesch2013state}. It will lead to wrong
orientation estimation. Second, the zero-velocity assumption used when deriving
\eqref{tro-eqn:baseline-measurement} is seldom true on hardware. The LO
velocity always underestimates the true velocity if the foot is rolling during
contact, as shown in Figure \ref{tro-fig:observation-model}. Both of these
limitations can be addressed by adding additional IMUs to the robot's feet. We
refer to this sensor-and-algorithm combination as Multi-IMU Proprioceptive
Odometry (Multi IMU PO). The hardware sensor setup will be explained in Section
\ref{tro-sec:experiments}, this section focuses on the algorithm. \rev{Compared
    to the previous publication \cite{yang2023multi}, we extend the filter state to
    include foot orientations and use a simplified quaternion small-rotation
    approximation technique to achieve fast and precise computation. We also add a
    new measurement model to keep the bias from drifting. }

We define the multi-IMU estimator state as $\bm{x} = [\bm{p}; \bm{v}; \bm{q};
    \bm{b}_a; \bm{b}_g; [\bm{s}; \dot{\bm{s}}; \bm{q}_f; \bm{b}_s; \bm{b}_t]_j]$
where $\bm{p}$, $\bm{v}$, $\bm{q}$, and $\bm{s}$ are the same as that defined
in Section \ref{tro-sec:sipo}. Additionally, we also estimate body IMU
accelerometer biases $\bm{b}_a \in \mathbb{R}^3$ and gyroscope biases $\bm{b}_g
    \in \mathbb{R}^3$, foot velocity $\dot{\bm{s}} \in \mathbb{R}^3$ and foot
orientation $\bm{q}_f \in SO(3)$ as a quaternion, and foot IMU accelerometer
biases $\bm{b}_s \in \mathbb{R}^3$ and foot IMU gyroscope biases $\bm{b}_q \in
    \mathbb{R}^3$. The subscript $j$ means that the second part of the state has
$E$ copies, one for each leg. \rev{The estimator state dimension increases
    considerably (for quadrupeds, the dimension is 80, while the single PO has a
    dimension of 28 even with biases added). But this dimension increase is
    necessary. As we will show, including foot orientation is important for keeping
    the entire system observable, and foot velocity is critical for low drift body
    position estimation. To keep a low computation load, we would like to calculate
    analytical Jacobians and leverage Jacobian sparsity as much as possible. }

For brevity, again in the subsequent discussion, we consider $E=1$ and drop leg
index $j$. We assume all sensors are synchronized and produce data at the same
frequency. In addition to sensor measurements introduced in Section
\ref{tro-sec:sipo}, we also get $\bm{a}_{f}$ and $\bm{\omega}_{f}$, the foot
acceleration reading and the foot angular velocity reading from an IMU
installed on the foot in the foot frame.

\subsection{EKF Process Model}
\rev{We use $\hat{\bm{x}}'$ to represent $\hat{\bm{x}}_{k+1|k}$ in order to simplify notation. The process model is changed from \eqref{tro-eqn:baseline-process} to }
\begin{align}
    \hat{\bm{x}}' =
    \begin{bmatrix}
        \hat{\bm{p}}'       \\
        \hat{\bm{v}}'       \\
        \hat{\bm{q}}'       \\
        \hat{\bm{b}}'_a     \\
        \hat{\bm{b}}'_g     \\
        \hat{\bm{s}}'       \\
        \dot{\hat{\bm{s}}}' \\
        \hat{\bm{q}}'_f     \\
        \hat{\bm{b}}'_s     \\
        \hat{\bm{b}}'_t     \\
    \end{bmatrix}=
    \begin{bmatrix}
        \hat{\bm{p}} + \Delta t\hat{\bm{v}}                                                  \\
        \hat{\bm{v}} + \Delta t(R(\hat{\bm{q}})(\bm{a}_b-\hat{\bm{b}}_a) - \bm{g}^w)         \\
        M(\hat{\bm{q}})\text{Exp}((\bm{\omega}_b-\hat{\bm{b}}_g)\Delta t)                    \\
        \hat{\bm{b}}_a                                                                       \\
        \hat{\bm{b}}_g                                                                       \\
        \hat{\bm{s}}+ \Delta t\dot{\hat{\bm{s}}}                                             \\
        \dot{\hat{\bm{s}}} + \Delta t(R(\hat{\bm{q}}_f)(\bm{a}_f-\hat{\bm{b}}_s) - \bm{g}^w) \\
        M(\hat{\bm{q}}_f)\text{Exp}((\bm{\omega}_f-\hat{\bm{b}}_t)\Delta t)                  \\
        \hat{\bm{b}}_s                                                                       \\
        \hat{\bm{b}}_t                                                                       \\
    \end{bmatrix}\label{tro-eqn:proposed-process}.
\end{align}
\rev{Compared to \eqref{tro-eqn:baseline-process}, although the dimension increases, the update equations are structurally identical and self-contained for body and foot, which are two rigid links with installed IMUs. Therefore, the Jacobian is fairly sparse. Moreover, the noise covariance of this process model does not depend on the contact flag because the dynamics are continuous. The foot velocity noise covariance is constant as long as the foot IMUs do not saturate their readings, which can be guaranteed by proper controller design and IMU hardware selection. }

\subsection{The Pivoting Contact Model}
\begin{figure}
    \setlength\fheight{0.7\linewidth}
    \setlength\fwidth{0.9\linewidth}
    \begin{minipage}[b]{0.3\linewidth}%
        \includegraphics[width=\linewidth]{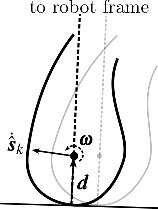}
        \vspace{7pt}
    \end{minipage}
    \begin{minipage}[b]{0.7\linewidth}
        \resizebox{\linewidth}{!}{%
            \input{tikz/foot_vel_compare}
        }
    \end{minipage}
    \caption{\textbf{Left:} Illustration of the pivoting model for a foot that has rolling contact with the ground. The estimated foot velocity $\dot{\bm{s}}_k$ depends on $\bm{\omega}$ and $\bm{d}$, as defined in \eqref{tro-eqn::omega} and \eqref{tro-eqn::r}. \textbf{Right:} Comparison of ground-truth foot velocity captured with a motion capture system with estimation using the pivoting model \eqref{tro-eqn::foot-v}. When the foot has nonzero rolling velocity during contacts (shaded regions), the pivoting model agrees with the ground truth velocity very well while the zero-velocity model treats foot velocity as zero.}
    \label{tro-fig:observation-model}
\end{figure}

\rev{A key observation that significantly enhances the estimation of body velocity is the relationship between the foot's contact point and its velocity. For any legged robot, whether it has point feet or flat feet, the foot pivots around the contact point at any given moment. This occurs irrespective of whether the contact foot is stationary or in motion, the type of terrain, or any deformation that might occur on the foot surface or the ground. Thus, the contact foot's linear velocity should equal the cross product of the foot angular velocity vector $\bm{\omega}$ and a pivoting vector $\bm{d}$ pointing from the contact point to the foot frame origin. For robots with flat feet, the contact point is at the center of pressure (CoP), which can be measured directly by foot contact sensors \cite{elnecavexavier:hal-04382871}. However, for robots equipped with spherical "point" feet, directly measuring the contact location on each foot is difficult. We discovered that a suitable approximation can be achieved by using the point on the foot's surface that aligns with a line drawn from the body frame origin to the foot frame origin. This approximation is effective because the body frame also pivots around the contact point.} This specific point and the corresponding equations are depicted in Figure \ref{tro-fig:observation-model}, in which
\begin{align}
    \dot{\hat{\bm{s}}}                           & = \bm{\omega}_p(\hat{\bm{x}}, \bm{\omega}_{f})\times \bm{d}_p(\hat{\bm{x}}, \bm{\alpha}), \text{and} \label{tro-eqn::foot-v} \\
    \bm{\omega}_p(\hat{\bm{x}}, \bm{\omega}_{f}) & = R(\hat{\bm{q}}_f)(\bm{\omega}_f-\hat{\bm{b}}_t) \label{tro-eqn::omega} ,                                                   \\
    \bm{d}_p(\hat{\bm{x}}, \bm{\alpha})          & = -d_0 \cdot \bm{n}/\|\bm{n}\|  .   \label{tro-eqn::r}
\end{align}
In the calculation, $d_0$ is the distance between the foot center and the foot surface. \rev{If the foot does not deform much during locomotion, this distance can be treated as a constant and measured from the CAD model, which is the case in our experiments; if the foot deforms a lot, $d_0$ can be obtained using data-driven methods or a calibration method such as that described in \cite{yang2022online}.}  $\bm{n} = R(\hat{\bm{q}})\bm{g}(\bm{\alpha})$ is the contact normal vector expressed in the world frame. This ``pivoting model'' captures the rolling contact better than \cite{yang2022online}.

\subsection{EKF Measurement Model}\label{tro-sec::measurement-model}

\rev{We define a measurement residual $y(\hat{\bm{x}}, \bm{\alpha},\bm{a}_{f},  \bm{\omega}_{f}) = \bar{\bm{z}}(\bm{\alpha}) -\bm{h}(\hat{\bm{x}}, \bm{\alpha},\bm{a}_{f},  \bm{\omega}_{f})$, where}
\begin{align}
    \bar{\bm{z}}(\bm{\alpha}) = \begin{bmatrix}
                                    \bm{g}(\bm{\alpha}) ;
                                    \bm{0} ;
                                    \bm{0} ;
                                    \bm{0}  ;
                                    \bm{a}_{f}
                                \end{bmatrix}\label{tro-eqn:mipo-measurement}
\end{align}
and
\begin{align}
     & \bm{h}(\hat{\bm{x}}, \bm{\alpha},\bm{a}_{f},  \bm{\omega}_{f}) = \\ &\begin{bmatrix}
        R(\hat{\bm{q}})^T(\hat{\bm{s}}-\hat{\bm{p}})                                                                                                     \\
        \text{Log}(\bm{q}(\bm{\alpha})^{-1}\otimes \hat{\bm{q}}^{-1}\otimes \hat{\bm{q}}_f)                                                              \\
        J(\bm{\alpha})\dot{\bm{\alpha}}-\skewcross{\bm{\omega}_b-\hat{\bm{b}}_g}\bm{g}(\bm{\alpha}) - R(\hat{\bm{q}})^T(\hat{\bm{v}}-\dot{\hat{\bm{s}}}) \\
        \dot{\hat{\bm{s}}} - \bm{\omega}_p(\hat{\bm{x}}, \bm{\omega}_{f})\times \bm{d}_p(\hat{\bm{x}}, \bm{\alpha})                                      \\
        R(\hat{\bm{q}}_f)^T\bm{g}^w+\hat{\bm{b}}_s
    \end{bmatrix}.\label{tro-eqn:mipo-residual}
\end{align}
\rev{Several terms in the sensor measurement vector $\bar{\bm{z}}$ are $\bm{0}$ and the sensor data appears in the measurement model. We call these ``implicit measurements.''} 

\rev{The first term of the residual $y$ is the same as in \eqref{tro-eqn:baseline-residual}. The second term uses forward kinematics to observe the rotational difference between the foot orientation and the body orientation, which is an implicit measurement because of the special structure of the quaternion. The term $\bm{q}(\bm{\alpha})$ represents a rotation between the foot frame and the body frame calculated from the forward kinematics \cite{lynch2017modern}. The third term comes from \eqref{tro-eqn:lo} without assuming $\bm{v}^w_f = 0$. In contrast to \eqref{tro-eqn:baseline-residual}, these three terms related to leg forward kinematics do not have varying measurement noise since they stay valid across foot contact switches. }

We refer to the fourth term, which is based on the pivoting model, as a
pivoting measurement. The pivoting measurement is only valid when a foot is in
contact with the ground and its residual value is drastically different in
stance and swing phases, which allows us to use a better technique to change
its contribution to the estimation.

\rev{The last term commonly appears in the IMU complementary filter~\cite{valenti2015keeping, mahony2008nonlinear}, where the IMU accelerometer reading is used for observing the gravity direction. This measurement model assumes the gravitational field dominates the low frequency response of the accelerometer. This assumption is often not valid for IMUs on the body of legged robots where horizontal linear accelerations are significant. However, when IMUs are on their feet and contact can be reliably detected, this measurement model becomes accurate and can contribute to robot orientation estimation significantly.  }

\rev{Since the yaw orientation of the robot is never observable \cite{bloesch2013state}, whenever a better yaw observation is available --- for example, a vision-based method from VIO where global landmarks can be observed --- we can introduce it into the EKF. It can be formulated in different ways, one of the simplest observations is calculating the yaw angle from the orientation quaternion as \cite{altmann2005rotations}}
\begin{equation}
    \text{YAW}(\bm{q}) = \arctan(2(q_x q_y + q_w q_z),1- 2 (q_yq_y + q_z q_z))
\end{equation}


\subsection{Foot Contact and Slip Detection}\label{tro-sec:mipo-outlier}
The last two terms in the measurement model~\eqref{tro-eqn:mipo-residual} are
only accurate when the robot foot has contact with the ground. Instead of doing
the covariance-scaling heuristic \eqref{tro-eqn:covariance-enlarge} for them,
we use a statistical test based on the Mahalanobis norm \cite{bloesch2013slip}:
\begin{equation}
    \|\bm{y}\|_{S} < \sigma,
    \label{tro-eqn:mahalanobis-test}
\end{equation}
where $\sigma$ is a hyperparameter, $\bm{y} = \dot{\hat{\bm{s}}} - \bm{\omega}_p\times \bm{d}_p$ as defined in \eqref{tro-eqn:mipo-residual}, and $S$ is its corresponding covariance matrix calculated in the Kalman filter step \eqref{tro-eqn:ekf-S}. If \eqref{tro-eqn:mahalanobis-test} is satisfied, we treat the foot as being in non-slipping contact and include the corresponding pivoting measurement in the update \eqref{tro-eqn:mipo-residual}.

Due to substantial differences in foot velocity between swing and stance phases
(see Figure \ref{tro-fig:observation-model} unshaded regions),
\eqref{tro-eqn:mahalanobis-test} can effectively distinguish foot phases
without a dedicated contact sensor. Notably, analogous mechanisms for the
zero-velocity model have been employed in previous works such as
\cite{bloesch2013state} and \cite{kim2021legged}.
In Section \ref{tro-sec:experiments}, we further demonstrate the importance of
the statistical test in the pivoting model through an ablation study.

\subsection{\rev{Analytical Jacobian}}\label{tro-sec:mipo-analytical-jacobian}
\rev{
It is well known that when the EKF state includes quaternions, the filter should operate on the error state \cite{madyastha2011extended, yang2022online}, in which orientation error is parameterized by a three-parameter rotation representation. Let the true state of the robot be $\bm{x}$ and the estimated state be $\hat{\bm{x}}$, then we denote the error state as  $\delta \bm{x} = [\delta \bm{p}; \delta \bm{v}; \delta \bm{\theta}; \delta \bm{b}_a; \delta \bm{b}_g; \delta \bm{s}; \delta \dot{\bm{s}}; \delta \bm{\theta}_f; \delta \bm{b}_s; \delta \bm{b}_t]$, in which Euclidean vectors are in the form of true state minus estimated state, such as $\delta \bm{p} = \bm{p} - \hat{\bm{p}}$. But the error rotation is $\delta \bm{\theta} = \text{Log}(\hat{\bm{q}}^{-1}\otimes\bm{q}) \in \mathbb{R}^3$ as discussed in Section \ref{tro-sec:small-angle}. In this section, we give the analytical form of the Multi-IMU PO's EKF model Jacobians $F$ and $H$ introduced in Section~\ref{tro-sec:ekf}, with detailed derivations of key terms in the Appendix.
}

\rev{
    The process Jacobian $F$ is a sparse block matrix }
\begin{align}
     & F(\bm{x}, \bm{a}, \bm{\omega}, \bm{a}_{f}, \bm{\omega}_{f}) = \label{tro-eqn:jacobian-process}                        \\
     & \begin{bmatrix}
           F_s(\bm{q},\bm{b}_a, \bm{b}_g, \bm{a}, \bm{\omega}) & 0                                                               \\
           0                                                   & F_s(\bm{q}_{f},\bm{b}_s, \bm{b}_t, \bm{a}_{f}, \bm{\omega}_{f})
       \end{bmatrix}\nonumber
\end{align}
\rev{
    where }
\begin{align}
     & F_s(\bm{q},\bm{b}_a, \bm{b}_g, \bm{a}, \bm{\omega}) =                                                                                               \\
     & \hspace{-11pt}\begin{bmatrix}
                         I & I\Delta t & 0                                                & 0          & 0                   \\
                         0 & I         & - R(\bm{q})\skewcross{(\bm{a}-\bm{b}_a)\Delta t} & 0          & - R(\bm{q})\Delta t \\
                         0 & 0         & I- \skewcross{(\bm{\omega}-\bm{b}_g)\Delta t}    & -I\Delta t & 0                   \\
                         0 & 0         & 0                                                & I          & 0                   \\
                         0 & 0         & 0                                                & 0          & I                   \\
                     \end{bmatrix}.\nonumber
\end{align}

\rev{
    We write the measurement Jacobian $H$ as}
\begin{equation}
    H(\bm{x}, \bm{\alpha},\bm{a}_{f},  \bm{\omega}_{f}) = [H_b \ \ H_f], \label{tro-eqn:jacobian-measurement}
\end{equation}
\rev{
    where the first block $H_b$ is Jacobian concerning robot body-related state variables while the second block relates to foot $H_f$. Let $R$ be short for $R(\bm{q})$ and $R_f=R(\bm{q}_f)$, then we have:}
\begin{align}
     & H_b =                                                                                                                                                                                                   \\
     & \hspace{-11pt} \begin{bmatrix}
                          -R^T & 0    & \skewcross{R^{T}(\bm{s}-\bm{p})}                                         & 0 & 0                           \\
                          0    & 0    & -\mathcal{L}(\bm{q}_f^{-1}\otimes \bm{q})\mathcal{R}(q(\bm{\alpha}))_{3} & 0 & 0                           \\
                          0    & -R^T & \skewcross{R^{T}(\dot{\bm{s}}-\bm{v})}                                   & 0 & -\skewcross{g(\bm{\alpha})} \\
                          0    & 0    & \skewcross{\bm{\omega}_p}\skewcross{\bm{d}_p}                            & 0 & 0                           \\
                          0    & 0    & 0                                                                        & 0 & 0
                      \end{bmatrix}\nonumber
\end{align}
\begin{align}
     & \text{and}\ H_f =                                                                                                                                                          \\
     & \hspace{-11pt} \begin{bmatrix}
                          R^T & 0   & 0                                                                  & 0 & 0                             \\
                          0   & 0   & \mathcal{L}(q(\bm{\alpha})\otimes \bm{q}^{-1}\otimes \bm{q}_f)_{3} & 0 & 0                             \\
                          0   & R^T & 0                                                                  & 0 & 0                             \\
                          0   & I   & -\skewcross{\bm{d}_p}\skewcross{\bm{\omega}_p}                     & 0 & -\skewcross{\bm{\omega}_p}R_f \\
                          0   & 0   & \skewcross{R_f^T\bm{g}^w}                                          & I & 0
                      \end{bmatrix}. \nonumber
\end{align}
\rev{
    In above equations, $\mathcal{L}()_3$ means the bottom right 3-by-3 sub-matrix of Equation \eqref{tro-eqn:left-quat-map} and $\mathcal{R}()_3$ is the bottom right 3-by-3 sub-matrix of Equation \eqref{tro-eqn:right-quat-map}. These Jacobians are sparse by nature. And their elements can be calculated efficiently. }

\rev{
    In the literature, some formulations also treat joint-angle and joint-angle velocity measurements as noise corrupted and calculate a measurement Jacobian for $\bm{\alpha}$ and $\dot{\bm{\alpha}}$ \cite{hartley2018contact, wisth2020preintegrated}. We found this Jacobian can often be approximated as a constant matrix since, during locomotion, joint angles stay within small ranges so we can precalculate the Jacobian using the average joint angles as a part of $\Sigma_w$.
}

\subsection{\rev{Observability Analysis}}
\rev{
    The nonlinear observability analysis of the dynamical system we have formulated is nontrivial because of the complicated measurement model. It is easier to do local observability analysis through the Jacobians. Prior work has pointed out that such analysis would lead to the conclusion that the yaw angle of the orientation is observable \cite{bloesch2013state}, which it is not. Intuitively, it is obvious that, even with additional IMUs, the global positions and yaw angles of the robot are still not observable. Therefore, our goal for observability analysis is to prove other state terms are observable.
}
\rev{
    We can verify that for the process Jacobian \eqref{tro-eqn:jacobian-process} and the measurement Jacobian \eqref{tro-eqn:jacobian-measurement}, the Observability Matrix \cite{huang2010observability}
}
\begin{equation}
    \mathcal{O}_1 =
    \begin{bmatrix}
        H_0                     \\
        H_1F_0                  \\
        \vdots                  \\

        H_TF_TF_{T-1}\cdots F_0 \\
    \end{bmatrix}
\end{equation}
\rev{
    and the local Observability Gramian \cite{krener2009measures}
}
\begin{equation}
    \mathcal{O}_2 =
    \sum_{i=0}^T (F_i^{T})^iH_i^TH_i(F_i)^i
\end{equation}
\rev{
    both have rank $15\times(1+E)-3$,
    where the Jacobians are evaluated along an estimated state trajectory $[x_0, x_1,\dots, x_T]$ with enough time steps. $15$ is the dimension of state variables related to either the robot body link or a foot link. $E$ is the total number of legs. This result aligns with the conclusion that global position is not observable and the yaw angle is rendered observable locally. Since the most important job of the state estimator is to provide pitch and roll orientation as well as body pose rate (linear velocity and angular velocity) to the controller, be sure that these key variables are observable is enough to ensure overall system stability.
}

\rev{
    The rank condition is satisfied for any state and measurement value, even when the robot stands still with zero angular velocity. The standard PO's observability matrix would drop rank when body angular velocity is $\bm{0}$, while the Multi-IMU PO will keep the rank condition consistent.
}

\subsection{Cram\'{e}r-Rao Lower Bound}\label{tro-sec:crlb} The Cram\'{e}r-Rao Lower Bound (CRLB) is
a metric that indicates the theoretical lower bound of the covariance of the
output of a state estimator \cite{taylor1979cramer}.
By computing and contrasting the CRLBs of various filters, we can validate our
hypothesis that multi-IMU PO is expected to outperform the conventional PO
method due to a richer set of sensor information. Even if in both cases the
position is not observable, the relative scale and growing rate of position
covariance can be used as a good indicator.

According to \cite{taylor1979cramer}, \rev{the CRLB for a Kalman filter is
    defined as a matrix $P^*_k$ at any time instance k. This matrix is computed
    recursively along with the EKF procedure as:}
\begin{equation}\label{tro-eqn:crlb}
    (P^*_k)^{-1} = (F_k P^*_{k-1} F_k^T)^{-1} + H_k^T\Sigma_w^{-1}H_k
\end{equation}
\rev{starting from $P^*_0 = P_0$. It is the lower bound on the estimation covariance $P_k$ such that $P_k \geq P^*_k$.}
$F_k$, $H_k$, and $\Sigma_w$ are defined in Section \ref{tro-sec:mipo-analytical-jacobian}. These matrices will have different values for Multi-IMU PO and the standard PO, but since the states of both filters contain the position of the robot, we can examine the covariance of position in their respective CRLBs, as will be shown in Section \ref{tro-sec:experiments}.

\section{Multi-IMU VILO}\label{tro-sec:cerberus2}
Building upon Multi-IMU PO, we present a Multi-IMU VILO state estimator
formulation that fuses camera images, body IMU, joint encoders, and foot IMUs
to achieve low-drift position estimation and precise orientation and velocity
estimation. This estimator follows the general structure of the
optimization-based VILO described in Section \ref{tro-sec:vilo}, and the
Multi-IMU PO output is used to formulate a motion constraint. \rev{More
    specifically, the velocity estimation from Multi-IMU PO is integrated into a
    contact preintegration constraint \cite{wisth2019robust}. A software
    implementation of this state estimator, Cerberus2 is built upon VILO software
    Cerberus \cite{yang2022cerberus}.}

\subsection{Software Architecture}
\begin{figure}
    \centering
    \includegraphics[width=\linewidth]{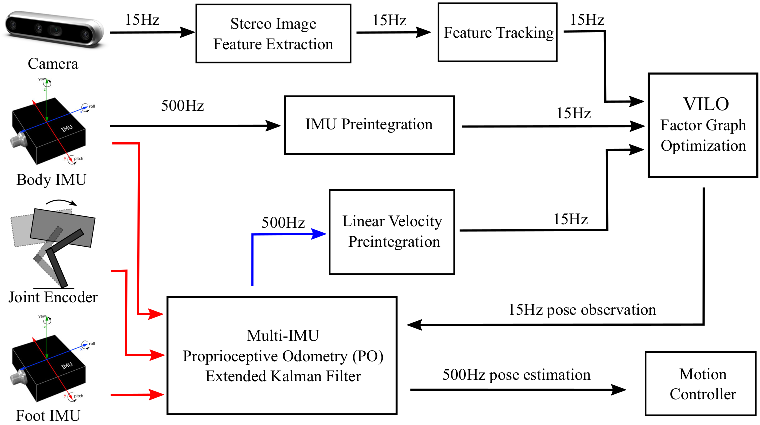}
    \caption{The software architecture of Cerberus2.}
    \label{tro-fig:sys_arch}
\end{figure}

\rev{Cerberus2 combines a VILO and the Multi-IMU PO in a loosely coupled way~\cite{wisth2019robust}}, as shown in Figure \ref{tro-fig:sys_arch}. In VILO, as Section \ref{tro-sec:vilo} explains, solving the nonlinear least square problem represented by a factor graph is a well-established technique. Using the marginalization to reduce factor graph size hence maintaining constant computation consumption is also very standard. It is how different factors are constructed that affects estimation accuracy the most.
The Multi-IMU PO interacts with the VILO in two ways. First, the VILO uses the velocity information generated by \eqref{tro-eqn:mahalanobis-test} in the Multi-IMU PO. Second, Multi-IMU PO corrects its yaw angle using the VILO output orientation estimation.

\subsection{PO Velocity Preintegration Constraint}


During the period between two VILO states $\hat{\bm{x}}_k$ and
$\hat{\bm{x}}_{k+1}$, there could be $L$ Multi-IMU PO measurements. Let $\nu_i$
be the PO velocity estimation at time $t_i$, and the associated covariance of
which is $\sigma_i$. The covariance can be extracted from the estimation
covariance of the PO output. A preintegration term is defined as:
\begin{align}
    \hat{\bm{\epsilon}}^{k}_{k+1} = \sum_{i=1}^L \nu_i \delta t \label{tro-eqn:loose-contact}
\end{align}
This term is the total body displacement measured by the leg odometry. It is in the world frame, so a residual is defined as:
\begin{align}
     & \bm{r}''(\hat{\bm{x}}_k, \hat{\bm{x}}_{k+1},Z_{\Delta k}) =  \begin{bmatrix}
                                                                        (\hat{\bm{p}}_{k+1} - \hat{\bm{p}}_{k}) - \hat{\bm{\epsilon}}^{k}_{k+1} \\
                                                                    \end{bmatrix}, \label{tro-eqn:loose-lo-residual}
\end{align}
The residual covariance for it is simply:
\begin{align}
    \hat{P}^{k}_{k+1} = \sum_{i=1}^L \sigma_i\delta t
\end{align}

\section{Implementation \& Experiments}\label{tro-sec:experiments}

We conducted extensive hardware experiments to evaluate the Multi-IMU PO and
the Multi-IMU VILO. We first describe the sensor hardware implementation and
experimental setup. Then, we evaluate the Multi-IMU PO component individually,
highlighting its significant performance gain over standard PO. Third, we
present the results of various indoor and outdoor experiments for Cerberus2 to
examine its robustness. Also, we study how different gait types, gait
frequencies, locomotion speeds, and terrain types affect the estimator
accuracy.

\subsection{Sensor Hardware}
We built the necessary sensor hardware using off-the-shelf low-cost components.
A Unitree Go 1 robot is selected as the locomotion platform, but the state
estimator should work on any quadruped, bipedal, and humanoid robot system.


The Cerberus2 hardware does not significantly alter the form factor of the Go1
robot. Four MPU9250 IMUs are installed on each foot. Since it is difficult to
directly install an IMU at the exact foot center, we transform foot IMU outputs
to the foot center frame using the transformation measured in the CAD model. A
Teensy 4.1 board communicates with the foot IMUs, acquiring their outputs at
200Hz. The Go1 robot's built-in proprietary MEMS IMU sensor and joint motor
encoder data can be obtained at 500Hz via Ethernet. The robot also has four
pressure contact sensors on the feet that can be thresholded to obtain binary
contact flags for the baseline method. Cerberus2, however, does not use contact
sensor data. We also installed an Intel Realsense camera on the robot to
provide 15Hz stereo vision RGB images. An Intel NUC computer with an i7-1165G7
CPU finally collects all sensor data from the Go1 robot, camera, and the Teensy
board to run the Cerberus2 algorithm. The computer also runs a nonlinear
predictive control locomotion controller \cite{OCS2} which has a number of
locomotion gaits available. The 500Hz feedback of the controller comes from the
Multi-IMU PO component of the Cerberus2.

\subsection{Evaluation Metrics}
\rev{It is worth describing the performance metric we are going to use in the experiments in detail first. Two widely used evaluation metrics adopted in the state estimation community are Relative pose error (RPE) and Absolute trajectory error (ATE)~\cite{sturm2012benchmark}, we make use of them when global misalignment exists. However, we also want to do a direct comparison of the estimated trajectory and the ground truth trajectory without doing any scaling or rotation to the estimation.  }

We use drift percentage to determine how much the estimated position deviates
from the actual position over the course of long-distance traveling. We also
note that the position to be compared only contains XY directions, for absolute
Z position can be readily measured by fusing barometer pressure data. Along the
travel trajectory, at each time instance $t$ we can calculate the total travel
distance $s$ from the ground truth data. Also, we denote the estimated position
as $\hat{\bm{p}}$ and the ground truth position as $\bm{p}$. Then the drift
percentage is: $$ DRIFT(t) = \frac{\|\bm{p}-\hat{\bm{p}}\|}{s} \times 100 \% $$
where $\|\bm{p}-\hat{\bm{p}}\|$ is the Root-Square-Error (RSE) of the position.
Since odometry often gradually accumulates error drifts over traveling, this
metric can better eliminate the effect of travel distance on the performance
evaluation.

When the robot moves in simple trajectories such as a line or a single loop,
then the final drift percentage $DRIFT(T)$, where $T$ is the final trajectory
time, is already informative. If the motion trajectory is more complicated, the
average drift percentage: $$ AVR\_DRIFT = \frac{1}{N}\sum_{t=t_0}^T DRIFT(t) $$
captures the drift performance over time better, where $N$ is the total number
of time instances. Similarly, the median drift or $MED\_DRIFT$ can be
calculated following the standard definition.

\begin{figure}
    \setlength\fheight{0.5\linewidth}
    \setlength\fwidth{1.2\linewidth}
    \centering
    \resizebox{1.0\linewidth}{!}{%
        \input{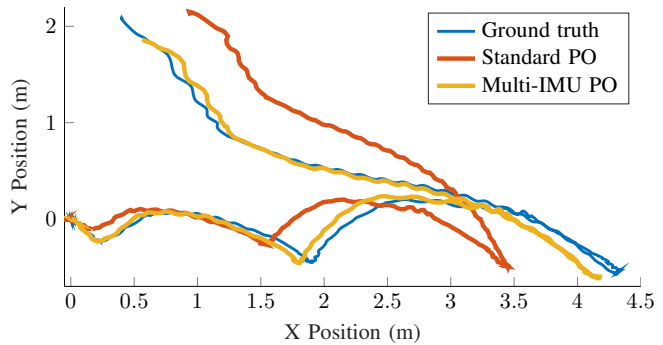}
    }
    \caption{The estimated XY position trajectory comparison of the standard PO and Multi-IMU PO. The total trajectory length is 10.5m. The trajectory of the standard PO drifts 11.39\% on average, and its maximum RSE is 1.04m. The position estimation results of the Multi-IMU PO has 2.31\% average drift and 0.25m maximum RSE. }
    \label{tro-fig:flying-trot}
\end{figure}
\subsection{Multi-IMU PO Position Estimation}\label{tro-sec:mipo-eval}
The Multi-IMU PO can be used individually for pose and velocity estimation. In
this section, we focus on its performance evaluation for position estimation.

We first compare the Multi-IMU PO against the standard PO in an indoor
environment. The robot operates in a lab space equipped with a motion-capture
(MoCap) system which provides the ground truth pose. The robot uses the trot
gait or flying trot gait (with a full airborne phase between leg switching, the
gait pattern is shown in Figure \ref{tro-fig:robustness-plot-gait}) to locomote
in arbitrary directions with a speed of 0.4-1.0 m/s on flat ground. To fairly
compare position estimation, for both the Multi-IMU PO and the standard PO, we
hold IMU biases as constant values and remove them from filter states because
experiment runs in this section are all within 1 minute. Also, we use the MoCap
ground truth orientation to directly form a measurement model for both filters
with the same noise parameters, so their orientations, especially yaw angles,
are drift-free.

\begin{table}[]
    \centering
    \caption{PO Performance Comparison}
    \begin{tabular}{|c|c|c|c|}
        \hline
                     & Frequency & Solve Time & Median Drift \\
        \hline
        Standard PO  & 500Hz     & 0.51ms     & 11.05\%      \\
        \hline
        Multi-IMU PO & 500Hz     & 1.69ms     & 2.61\%       \\
        \hline
    \end{tabular}
    \label{tro-tab:runtime}
\end{table}

Figure \ref{tro-fig:flying-trot} compares estimated position trajectories on
one data sequence. Table \ref{tro-tab:runtime} compares their per-loop solve
times with C++ implementation and average drifts across five different data
sequences.
Multi-IMU PO, despite a longer computation time (but still within the 500Hz
control requirement) due to larger state dimensions, significantly reduces
drift percentage compared to standard PO.



\begin{figure}
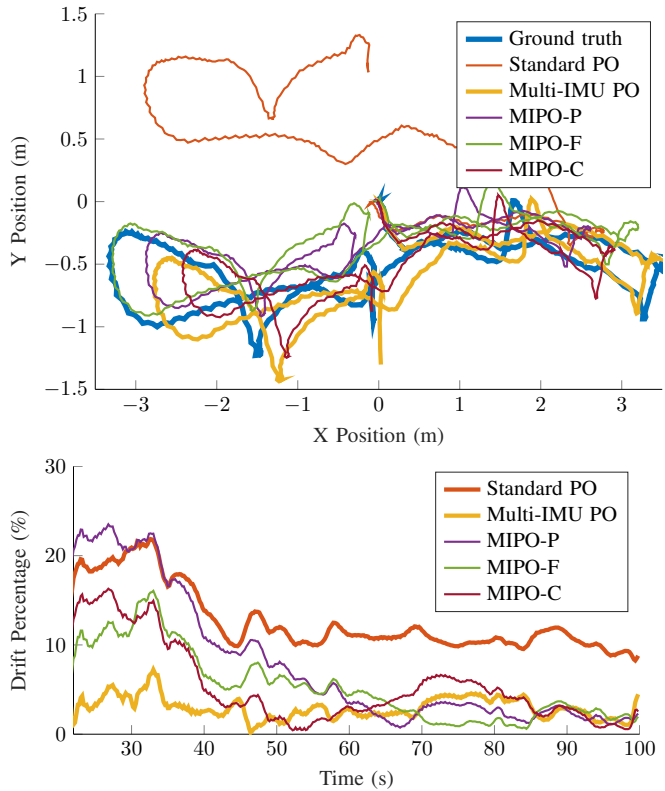

    \setlength\fheight{0.7\linewidth}
    \setlength\fheighta{0.5\linewidth}
    \setlength\fwidth{1.1\linewidth}
    \centering
    \resizebox{1.0\linewidth}{!}{%
        \input{tikz/mipo_ablation_traj2.tex}
    }
    \resizebox{1.0\linewidth}{!}{%
        \input{tikz/mipo_ablation_rse2.tex}
    }
    \caption{\textbf{Top:} XY position trajectory estimation results of the standard PO, Multi-IMU PO, and three Multi-IMU PO variants used in the ablation study. The total trajectory length is 21.5m. \textbf{Bottom:} The $DRIFT(t)$ of all methods. In terms of $AVR\_DRIFT$, Multi-IMU PO has the smallest (4.21\%), followed by MIPO-F (7.75\%), MIPO-C (11.3\%), MIPO-P (12.04\%), and the standard PO (16.87\%).}
    \label{tro-fig:ablation-compare}
\end{figure}


\subsection{\rev{Multi-IMU PO Orientation Estimation}}\label{tro-sec:mipo-eval-orientation}
\rev{Robot state estimation can be challenging in some special cases during locomotion. For example, when a legged robot performs fast in-place rotation with rapidly changing angular accelerations using the trotting gait, the robot could suffer from several error sources. First, robot feet deform more than usual, making leg kinematics less accurate. Second, the foot-supporting line gets close to roll axis so the roll angle becomes less observable. Lastly, angular acceleration makes IMU have elevated noise.  }

\begin{figure}
    \setlength\fheight{1.6\linewidth}
    \setlength\fwidth{1.1\linewidth}
    \centering
    \resizebox{1.0\linewidth}{!}{%
        \input{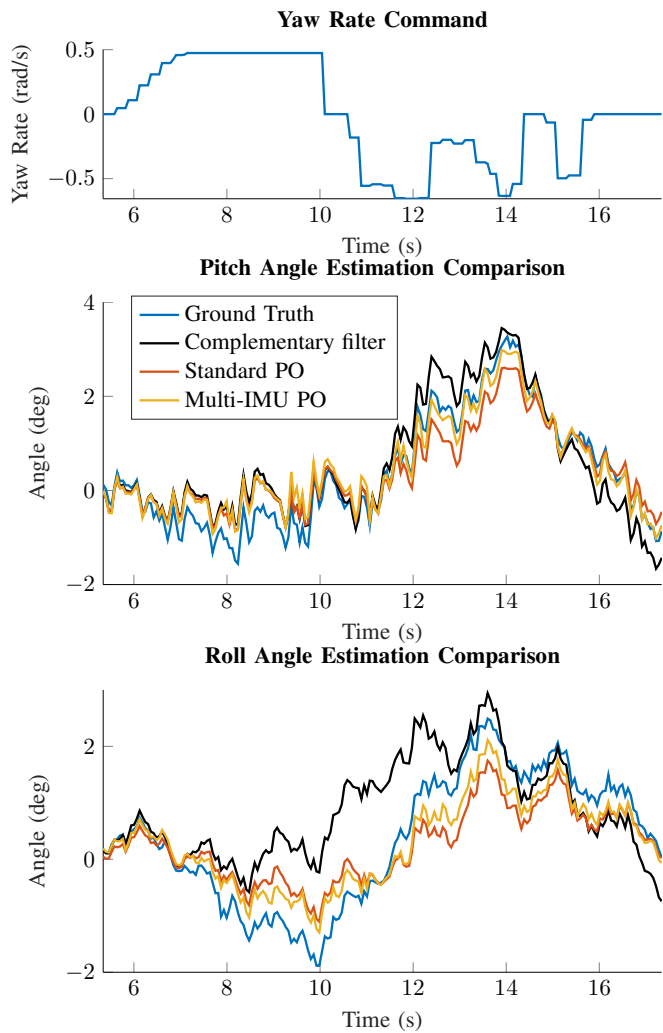}
    }
    \caption{
        \rev{To show how different orientation estimation methods perform differently during fast in-place rotation, we compare the user input yaw rate command to the robot (\textbf{Top}) and estimated body pitch (\textbf{Middle}) and roll angles (\textbf{Bottom}). Positive command lets the robot spin counterclockwise. The faster the robot spins and the more abrupt the angular acceleration is, the more the estimated orientation will be wrong. Compared with the MoCap ground truth, average pitch and roll angular estimation errors (in deg) of different methods are: Complementary filter (0.019, 0.040), Standard PO (0.0185, 0.025), Multi-IMU PO (\textbf{0.015}, \textbf{0.018}). }
    }
    \label{tro-fig:rotation-compare}
\end{figure}

\rev{These problems can be effectively solved by including foot orientations to the filter and adding gravity direction observation as shown in Equation~\ref{tro-eqn:mipo-measurement}. We observe that even during fast in-place rotation and foot-rolling contacts, the IMUs on feet capture gravity direction reasonably well during contacts. We show its contribution in Figure~\ref{tro-fig:rotation-compare}. In this experiment, the operator arbitrarily gave the robot a fast fast-changing yaw rate command to rotate the robot. We collect sensor data and offline run different estimation methods to get body orientation estimations. Unlike in the previous section, we do not use Mocap ground truth orientation to correct any estimation. Then we convert all estimated orientations to Euler Angle and compare pitch and roll angles with MoCap ground truth. In the plots, the ``Complementary Filter'' is an implementation of \cite{mahony2008nonlinear}, where the accelerometer and user command are used to estimate gravity direction to correct orientation drift. Among all methods, the Multi-IMU PO has the smallest orientation estimation error. In the most extreme case (time 13.6s, where the robot did a fast stop and start of rotation), ground truth roll orientation is 2.49deg, while standard PO estimates it to be 1.74deg and the Multi-IMU PO estimation is 2.11deg. The Multi-IMU's orientation is 50.6\% more accurate at this time and has 30\% less estimation error on average. }

\subsection{Multi-IMU PO Ablation Study}\label{tro-sec:mipo-ablation}

Next, through an ablation study, we investigate the individual contribution of
the pivoting model and the statistical test introduced in Section
\ref{tro-sec::measurement-model}. We create three algorithm variants: 1)
MIPO-P, where the last term in \eqref{tro-eqn:mipo-residual} is
$\dot{\hat{\bm{s}}}_k$ instead of $\dot{\hat{\bm{s}}}_k - \bm{\omega}\times
    \bm{d}$. We vary its measurement noise according to the contact flag but do not
perform the statistical test \eqref{tro-eqn:mahalanobis-test}. In this way,
this term is the same as the standard zero velocity foot assumption. 2) MIPO-F,
which is the same as the Multi-IMU PO. However, it replaces the statistical
test with the measurement noise adjustment mechanism based on the contact flag.
And 3) MIPO-C, which uses the Multi-IMU PO's process model, MIPO-P's
measurement model, and the statistical test on the pivoting model term instead
of varying measurement noise like MIPO-P. The results of standard PO, Multi-IMU
PO, and all three variants are shown in Figure \ref{tro-fig:ablation-compare}.

MIPO-F performs closer to Multi-IMU PO than standard PO, MIPO-P, and MIPO-C,
showing that the largest performance contributor is the pivoting constraint.
This result suggests that the zero-velocity model fundamentally limits the
capability of standard PO to achieve low-drift position estimation, even if
more accurate contact-flag generation methods can be used to avoid
contact-detection errors in standard PO.

\subsection{Multi-IMU CRLB}\label{tro-sec:mipo-exp-crlb}
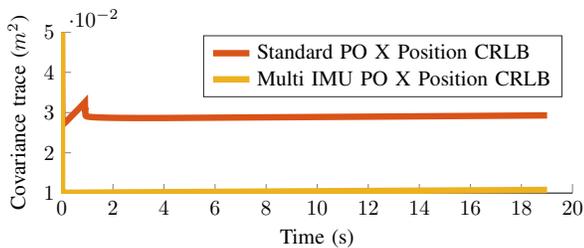
\begin{figure}
    \setlength\fheight{0.3\linewidth}
    \setlength\fwidth{1.0\linewidth}
    \centering
    \resizebox{1.0\linewidth}{!}{%
%
%
\definecolor{mycolor1}{rgb}{0.85098,0.32549,0.09804}%
\definecolor{mycolor2}{rgb}{0.92941,0.69412,0.12549}%

\begin{tikzpicture}

  \begin{axis}[%
      width=0.951\fwidth,
      height=\fheight,
      at={(0\fwidth,0\fheight)},
      scale only axis,
      xmin=0,
      xmax=20,
      xlabel={Time (s)},
      ymin=0.01,
      ymax=0.0500000000000007,
      ylabel={Covariance trace ($m^2$)},
      axis background/.style={fill=white},
      axis x line*=bottom,
      axis y line*=left,
      legend style={legend cell align=left, align=left, draw=white!15!black}
    ]
    \addplot [color=mycolor1, line width=3.0pt]
    table[row sep=crcr]{%
        0	0.0500000000000007\\
        0.00499999999999901	0.0375063429402083\\
        0.0100000000000016	0.0322164763009205\\
        0.0150000000000006	0.0296977756501917\\
        0.0199999999999996	0.0284337466570399\\
        0.0249999999999986	0.0277848187369329\\
        0.0300000000000011	0.0274502285604861\\
        0.0350000000000001	0.0272800562236348\\
        0.0399999999999991	0.0271971838590233\\
        0.0450000000000017	0.0271612513965067\\
        0.0549999999999997	0.0271548562102701\\
        0.0700000000000003	0.0272041132297787\\
        0.100000000000001	0.0273560920231297\\
        0.145	0.0276293562528949\\
        0.734999999999999	0.0314272377584821\\
        0.899999999999999	0.0325472526795991\\
        0.905000000000001	0.0306596375882293\\
        0.914999999999999	0.0306347347656093\\
        0.920000000000002	0.0294944567600481\\
        0.925000000000001	0.0293408432975752\\
        0.93	0.0292611922206056\\
        0.940000000000001	0.0291726919706896\\
        0.954999999999998	0.0291029885049277\\
        0.98	0.0290406650853541\\
        1.02	0.0289835997624657\\
        1.105	0.0289151583466207\\
        1.245	0.0288458850742686\\
        1.465	0.028777857908505\\
        1.79	0.0287184422524831\\
        2.275	0.0286716881405091\\
        2.965	0.0286458197714481\\
        4.04	0.02865578667598\\
        13.37	0.0290378263104287\\
        13.91	0.0290636498459236\\
        14.86	0.0291111721014978\\
        17.04	0.029208232916158\\
        19.005	0.0292979737055816\\
      };
    \addlegendentry{Standard PO X Position CRLB}

    \addplot [color=mycolor2, line width=3.0pt]
    table[row sep=crcr]{%
        0	0.0500000000000007\\
        0.00499999999999901	0.0115663462271343\\
        0.0100000000000016	0.0108059842252537\\
        0.0150000000000006	0.0105450105603886\\
        0.0199999999999996	0.0104130221140224\\
        0.0249999999999986	0.0103334489318563\\
        0.0350000000000001	0.0102425392910952\\
        0.0500000000000007	0.010175261925113\\
        0.0700000000000003	0.0101322815222638\\
        0.105	0.0101003147343732\\
        0.175000000000001	0.0100827980978409\\
        0.375	0.0100864827051446\\
        2.915	0.0101909300702161\\
        3.625	0.0102141684159278\\
        5.63	0.0102866954479737\\
        6.91	0.0103326470982488\\
        12.84	0.010549910813193\\
        14.485	0.0106105418453524\\
        16.09	0.0106702280073812\\
        19.005	0.0107768061948441\\
      };
    \addlegendentry{Multi IMU PO X Position CRLB}
  \end{axis}

  \begin{axis}[%
      width=1.227\fwidth,
      height=1.227\fheight,
      at={(-0.16\fwidth,-0.135\fheight)},
      scale only axis,
      xmin=0,
      xmax=1,
      ymin=0,
      ymax=1,
      axis line style={draw=none},
      ticks=none,
      axis x line*=bottom,
      axis y line*=left
    ]
  \end{axis}
\end{tikzpicture}%
    }
    \caption{Plots of the element value corresponding to X position in the CRLB matrix for standard PO and Multi-IMU PO.
        The CRLB of Multi-IMU PO is much smaller than that of standard PO.}
    \label{tro-fig:crlb}
\end{figure}

Lastly, we compare the CRLBs of Multi-IMU PO and standard PO. As discussed in
Section \ref{tro-sec:crlb}, the CRLB represents the lowest possible uncertainty
of a state estimator. For any given trajectory, we can calculate the CRLB
recursively as Equation \eqref{tro-eqn:crlb} from an initial covariance. In
Figure \ref{tro-fig:crlb}, it is shown that Multi-IMU PO has a much smaller
CRLB than standard PO. This result aligns with our intuition and previous
experiment results that Multi-IMU has better estimation performance.

\subsection{Cerberus2 Evaluation}
In this section, we evaluate the position estimation performance of Cerberus2.
In addition to comparing with the ground truth, for each data sequence, we also
make comparisons against baseline methods including Multi-IMU PO, VIO
(VINS-Fusion), and Single IMU VILO (Cerberus).




\subsubsection{Indoor Drift \& Robustness Evaluation}
\begin{figure}
    \setlength\fheight{0.6\linewidth}
    \setlength\fwidth{1.5\linewidth}
    \setlength\fheighta{0.3\linewidth}
    \setlength\fwidtha{1.5\linewidth}
    \centering
    \resizebox{1.05\linewidth}{!}{%
%
%
\begin{tikzpicture}

    \begin{axis}[%
            width=0.281\fwidtha,
            height=0.822\fheighta,
            at={(0\fwidtha,0\fheighta)},
            scale only axis,
            xmin=0,
            xmax=1,
            xtick={0,0.5,1},
            xticklabels={{0\%},{Gait Phase},{100\%}},
            ymin=0,
            ymax=2,
            ytick={0.25,0.75,1.25,1.75},
            yticklabels={{Leg1},{Leg2},{Leg3},{Leg4}},
            axis background/.style={fill=white},
            title style={font=\bfseries},
            title={Standing Trot},
            axis x line*=bottom,
            axis y line*=left
        ]
        \draw[fill=black, draw=white] (axis cs:0,0) rectangle (axis cs:0.55,0.5);
        \draw[fill=black, draw=white] (axis cs:0.5,0.5) rectangle (axis cs:1,1);
        \draw[fill=black, draw=white] (axis cs:0,1) rectangle (axis cs:0.55,1.5);
        \draw[fill=black, draw=white] (axis cs:0.5,1.5) rectangle (axis cs:1,2);
        \draw[fill=black, draw=white] (axis cs:0,0.5) rectangle (axis cs:0.05,1);
        \draw[fill=black, draw=white] (axis cs:0,1.5) rectangle (axis cs:0.05,2);
    \end{axis}

    \begin{axis}[%
            width=0.281\fwidtha,
            height=0.822\fheighta,
            at={(0.36\fwidtha,0\fheighta)},
            scale only axis,
            xmin=0,
            xmax=1,
            xtick={0,0.5,1},
            xticklabels={{0\%},{Gait Phase},{100\%}},
            ymin=0,
            ymax=2,
            ytick={0.25,0.75,1.25,1.75},
            yticklabels={{Leg1},{Leg2},{Leg3},{Leg4}},
            axis background/.style={fill=white},
            title style={font=\bfseries},
            title={Trot},
            axis x line*=bottom,
            axis y line*=left
        ]
        \draw[fill=black, draw=black] (axis cs:0,0) rectangle (axis cs:0.5,0.5);
        \draw[fill=black, draw=black] (axis cs:0.5,0.5) rectangle (axis cs:1,1);
        \draw[fill=black, draw=black] (axis cs:0,1) rectangle (axis cs:0.5,1.5);
        \draw[fill=black, draw=black] (axis cs:0.5,1.5) rectangle (axis cs:1,2);
    \end{axis}

    \begin{axis}[%
            width=0.281\fwidtha,
            height=0.822\fheighta,
            at={(0.719\fwidtha,0\fheighta)},
            scale only axis,
            xmin=0,
            xmax=1,
            xtick={0,0.5,1},
            xticklabels={{0\%},{Gait Phase},{100\%}},
            ymin=0,
            ymax=2,
            ytick={0.25,0.75,1.25,1.75},
            yticklabels={{Leg1},{Leg2},{Leg3},{Leg4}},
            axis background/.style={fill=white},
            title style={font=\bfseries},
            title={Flying Trot},
            axis x line*=bottom,
            axis y line*=left
        ]
        \draw[fill=black, draw=white] (axis cs:0,0) rectangle (axis cs:0.45,0.5);
        \draw[fill=black, draw=white] (axis cs:0.55,0.5) rectangle (axis cs:1.05,1);
        \draw[fill=black, draw=white] (axis cs:0,1) rectangle (axis cs:0.45,1.5);
        \draw[fill=black, draw=white] (axis cs:0.55,1.5) rectangle (axis cs:1.05,2);
    \end{axis}

    \begin{axis}[%
            width=1.127\fwidtha,
            height=1.127\fheighta,
            at={(-0.073\fwidtha,-0.147\fheighta)},
            scale only axis,
            xmin=0,
            xmax=1,
            ymin=0,
            ymax=1,
            axis line style={draw=none},
            ticks=none,
            axis x line*=bottom,
            axis y line*=left
        ]
    \end{axis}
\end{tikzpicture}%
    }
    \resizebox{1.0\linewidth}{!}{%
        \input{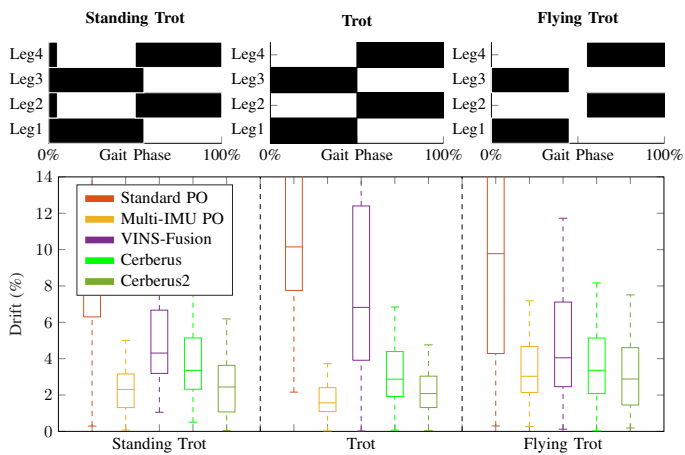}
    }
    \caption{\textbf{Top:} Leg contact schedules of each gait type. Black represents the periods where the leg is in contact with the ground. \textbf{Bottom:} Box plots of average drifts of different methods. Results are categorized by gait types. For each type, we show the drift statistics from all experiment runs for each method.}
    \label{tro-fig:robustness-plot-gait}
\end{figure}

We first did the same experiment as discussed in Section
\ref{tro-sec:mipo-eval}, but we want to understand more about the factors that
affect position estimation drifts.
We aim to understand how different locomotion conditions influence estimation
through a large number of controlled experiments. The three independent
locomotion-related variables we adjust in these experiments are gait type
(trot, standing trot, and flying trot), target moving speed (0.2$m/s$,
0.4$m/s$, 0.6$m/s$, 0.8$m/s$, 1.0$m/s$, and 1.2$m/s$), and gait time (0.32s,
0.36s, 0.40s, 0.44s, 0.48s, and 0.52s). The foot contact schedule of each gait
type is shown in Figure \ref{tro-fig:robustness-plot-gait}. The trot gait,
containing two alternating double support phases, is widely used in the
community as one of the most reliable gaits. The standing trot gait adds a 50ms
stance period after each trot foot switching, while the flying trot gait adds a
50ms full flight phase. The gait time is defined as the total duration of
double support phases. The moving speed is a target speed that the locomotion
controller wants to reach as soon as possible in each experiment run. A human
operator provides the speed target during experiments. For faster target linear
moving speed, the target yaw rate is also scaled up to ensure smooth turning.
We would like to emphasize that no prior work has studied the effect of this
wide range of moving speeds and gait variations on state estimation
performance.

57 experiment runs are conducted in the same lab space mentioned in Section \ref{tro-sec:mipo-eval}. In each run, we select a set of variable combinations for the robot controller, and then the robot travels a square trajectory that is used across experiments.
So during all experiment runs, the visual features along the trajectory are
roughly the same. Although the trajectories are short, the robot experiences
fast rotations, foot slippages, and unexpected contacts when moving at high
speed. Since long gait time will lead to controller instability at high moving
speeds, not all variable combinations are tested. We collect sensor data from
one run into one rosbag and run all baseline methods and Cerberus2 methods
using the sensor data offline to make a fair comparison. All methods use
consistent parameters even though controller parameters change so that we can
examine their parameter robustness. After obtaining the estimation results for
all runs, these results were then categorized into three groups for focused
analysis of each independent variable.

\begin{figure}
    \setlength\fheight{0.6\linewidth}
    \setlength\fwidth{1.5\linewidth}
    \centering
    \resizebox{1.0\linewidth}{!}{%
        \input{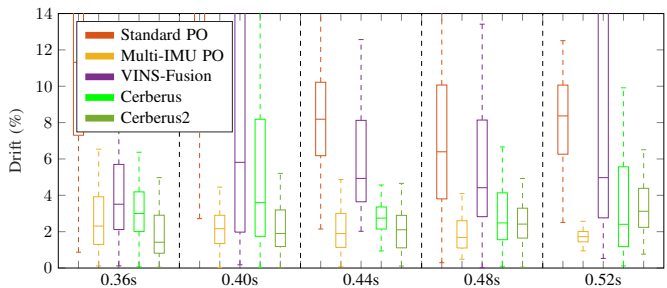}
    }
    \caption{Box plots for estimation results of different gait frequencies. }
    \label{tro-fig:robustness-plot-frequency}
\end{figure}
\begin{figure}
    \setlength\fheight{0.6\linewidth}
    \setlength\fwidth{1.5\linewidth}
    \centering
    \resizebox{1.0\linewidth}{!}{%
        \input{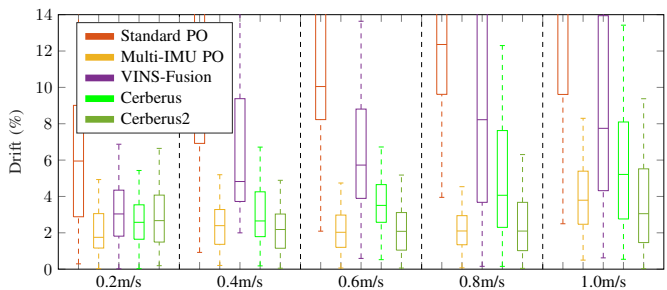}
    }
    \caption{Box plots for estimation results of different movement speeds. }
    \label{tro-fig:robustness-plot-speed}
\end{figure}





In Figure \ref{tro-fig:robustness-plot-gait} we compare the average drifts of
different gaits. Among the three gaits, the trot gait has the lowest drift and
the flying trot has the largest. Because of the airborne phase, the flying trot
makes the robot experience large impacts upon touchdown so the worst
performance is not surprising. Between the trot gait and the standing trot, the
trot has a shorter contact time and fewer rolling contacts, which could explain
its better performance. Therefore standard trot gait is preferred for achieving
the best position estimation performance.

In Figure \ref{tro-fig:robustness-plot-frequency}, we can see that as gait time
increases, the drift estimation results of the Multi-IMU PO monotonically
decrease while other vision-based methods do not show a clear changing pattern.
The VINS-Fusion even got much worse with a gait time of 0.52s. The reason is as
gait time increases, the robot will have larger body displacement per stride
hence larger joint angle velocity. According to Kalman filter observability
analysis \cite{yang2022online}, larger joint angle velocity will improve the
numerical condition of the Kalman gain. On the other hand, as long as the robot
controls its orientation well, gait time does not affect camera observation
accuracy. However, if the gait time is too long and the robot target velocity
is also high, the robot legs may exceed the workspace limit, resulting in
unstable locomotion, which leads to rapid body rotation and tampers the
cameras' imaging quality. Around 0.44s gait time, Multi-IMU PO (min drift
1.31\%, average drift 1.50\%) and Cerberus2 (min drift 1.08\%, average 2.07\%)
achieve the best drift performance.

As shown in Figure \ref{tro-fig:robustness-plot-speed}, when robot target speed
increases, VINS-Fusion's drift monotonically increases, so do Standard PO,
Cerberus, and Cerberus2. However, Multi-IMU PO and Cerberus2's drift
performances are always around 2\% until the target speed is higher than
$1m/s$. By looking at sensor data we notice at $1m/s$ some foot IMUs'
accelerometer readings occasionally exceed the range limit ($\pm 150m/s^2$).
Our filter implementation does not consider sensor saturation, therefore it
will diverge if saturation happens often. Another interesting phenomenon is in
the speed range $0.6m/s$-$0.8m/s$, Multi-IMU PO and Cerberus2 perform the best.
The reason could be faster speed leads to fewer foot contact switches along a
fixed travel distance, which reduces the effect of locomotion disturbances in
the same way as increasing gait time.

\begin{figure*}
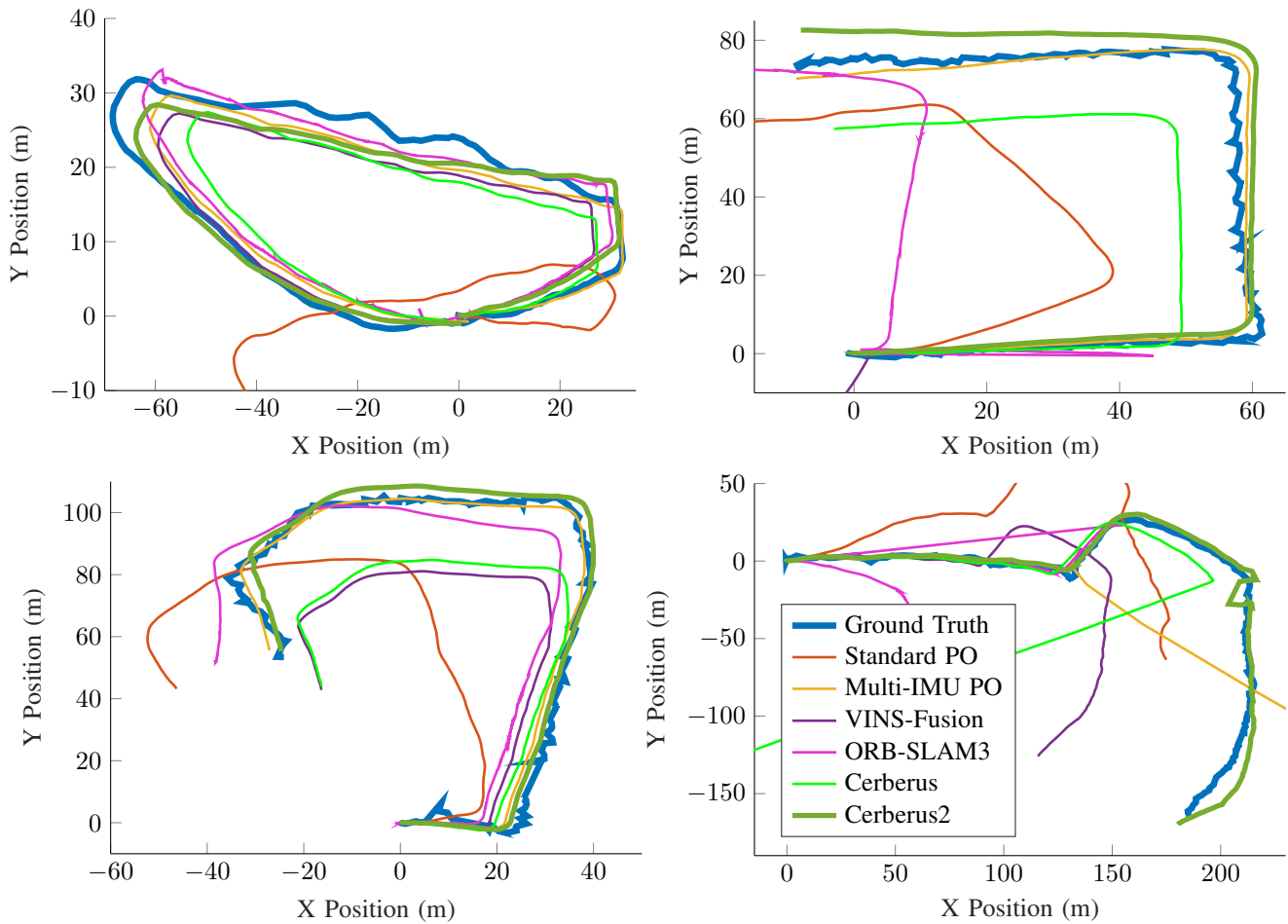

    \setlength\fheighta{0.4\linewidth}
    \setlength\fwidtha{0.4\linewidth}
    \newcommand\figurescale{0.6}
    \centering

    \hspace{-5pt}\input{tikz/outdoor_runs/tro_outdoor_frick_park_arch}
    \hspace{10pt}\input{tikz/outdoor_runs/tro_outdoor_mill19_park}
    \input{tikz/outdoor_runs/tro_outdoor_schenley_tennis}
    \hspace{-5pt}\input{tikz/outdoor_runs/tro_outdoor_st_mary}

    \caption{Estimation results of four outdoor runs. \rev{Quantitative analysis is summarized in Table \ref{tro-tab:outdoor-results}. In the bottom right run, the Multi-IMU PO breaks because the sudden impact on foot causes IMU saturation and wrong foot velocity measurement. When the Multi-IMU PO is used within Cerberus2, we can reset it using VILO's state so Cerberus2 can continue to run.}}
    \label{tro-fig:outdoor-4}
\end{figure*}

\subsubsection{Outdoor Drift \& Terrain Robustness Evaluation}
\begin{figure}
    \centering
    \includegraphics[width=\linewidth]{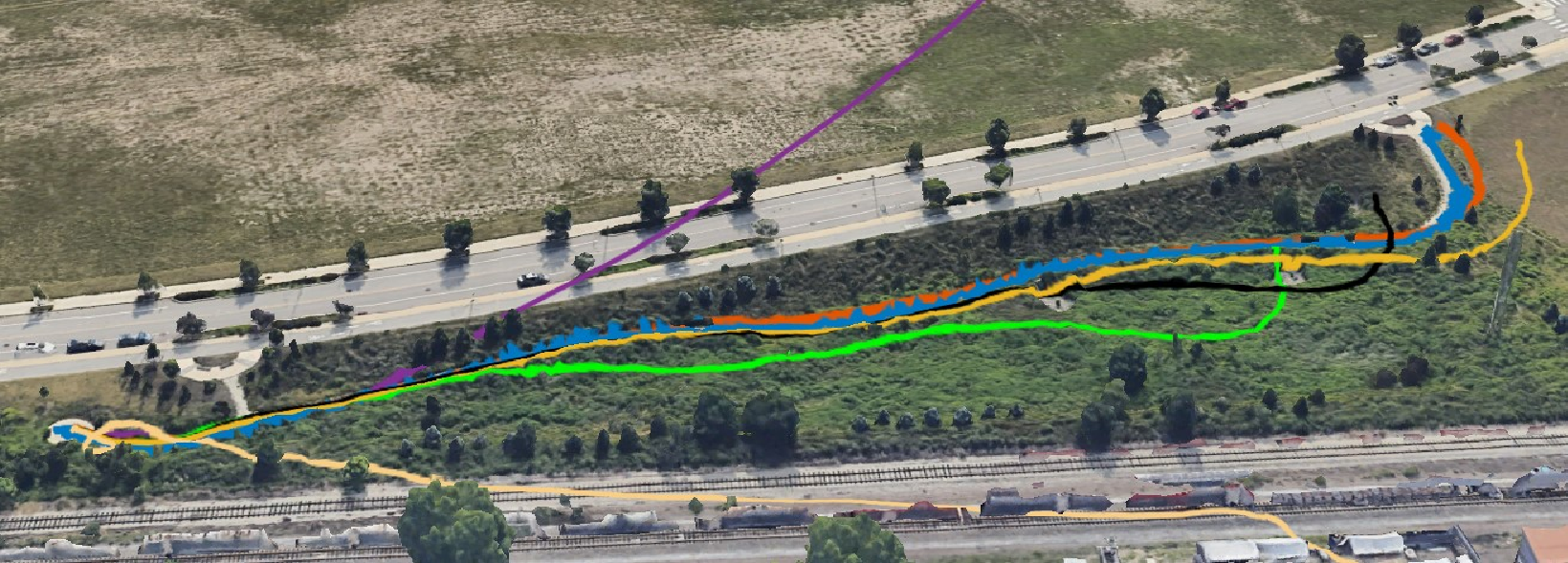}
    \caption{On a challenging hiking trail, Cerberus2 (red) achieves lower than 10\% drift performance while all other methods either fail or have drift larger than 10\%. The estimation result of Cerberus2 agrees with the ground truth (blue) and satellite image very well. The map can be viewed \href{https://www.google.com/maps/d/u/0/edit?mid=1ek1hU02ADSrNKmgSpMtXkDp6IPvOf6E\&ll=40.41327556703216\%2C-79.95071363609686\&z=19}{online}.}
    \label{tro-fig:outdoor-trail}
\end{figure}

To fully test the accuracy, adaptability, and robustness of Cerberus2, we let
the robot travel over uneven terrains for long distances
. A dataset that includes 9 experiment runs with a total length of over 2.5km is collected. During each run, in addition to collecting sensor data, we also mounted an iPhone on the robot, from which we measured a separate sequence of GPS and IMU data from the cellphone. \rev{Then we use MATLAB's built-in inertial filter \texttt{insfilterAsync}\footnote{\href{https://www.mathworks.com/help/nav/ref/insfilterasync.html}{https://www.mathworks.com/help/nav/ref/insfilterasync.html}} to recover the ground truth.} Ground truth trajectories are less accurate when GPS quality is poor, but in normal situations, they agree with satellite images very well, as shown in Figure~\ref{tro-fig:outdoor-trail}.

For each sequence, we compare Cerberus2 with VINS-Fusion, Cerberus, and
ORB-SLAM3 \cite{campos2021rob3}, an open-source localization library that has a
multi-map loop closure ability that should improve estimation accuracy. In
Figure \ref{tro-fig:outdoor-4}, we visualize the estimation results of 4
outdoor runs.

Figure \ref{tro-fig:outdoor-garage} demonstrates the robustness of Cerberus2
during two indoor/outdoor switchings. In Figure \ref{tro-fig:outdoor-trail},
estimation results for the robot traveling along a hiking trail are shown.
Because of gravel and plants on the trail surface, the robot often experiences
foot slippage and unexpected contact.

Since our robot sensor is low cost, robot movement is agile, and experiment
runs involve large-scale terrain variations, baseline vision-based methods, and
PO methods often fail or have large drift. Among all methods and across all
experiments, Cerberus2 has the best performance.

\subsubsection{Run Time}
Cerberus2 can run in real-time on robot hardware with moderate computation
resources. In our implementation, Cerberus2 takes less than 60ms to finish a
single factor graph solve on the Intel i7-1165G7 CPU, which is shorter than the
camera image arrival interval (1/15s or 66.7ms).
More implementation details and run time tips can be viewed in our open-source
codebase.

\begin{table}[]
    \centering
    \begin{tabular}{|c|c|c|c|c|}
        \hline
                           & Top-left        & Top-right       & Bottom-left     & Bottom-right    \\
        \hline
        Standard PO        & 17.67\%         & 19.62\%         & 14.15\%         & 26.24\%         \\
        \hline
        Multi-IMU PO       & 2.70\%          & \textbf{1.11\%} & \textbf{1.07\%} & break           \\
        \hline
        VINS-Fusion        & 3.04\%          & break           & 9.75\%          & 19.61           \\
        \hline
        ORB-SLAM3          & 1.83\%          & break           & 5.71\%          & break           \\
        \hline
        Cerberus           & 4.95\%          & 8.41\%          & 8.57\%          & break           \\
        \hline
        \textbf{Cerberus2} & \textbf{1.76\%} & 2.84\%          & 2.04\%          & \textbf{2.73\%} \\
        \hline
    \end{tabular}
    \caption{The drift value of all outdoor experiment runs. Each column lists the results of one run. Across all results, Multi-IMU PO achieves the lowest drift values, while Cerberus2 has comparable or better performance and improved robustness.}
    \label{tro-tab:outdoor-results}
\end{table}


\section{Limitation \& Future Work}\label{tro-sec:discussion}

The biggest limitation of Multi-IMU PO and VILO is that we do not have the
capability of dealing with sensor saturation. This is the primary reason for
estimator failures if the robot motion controller is not well designed.
Incorporating some signal reconstruction or saturation detection mechanism to
further improve robustness is a promising future work.

In VILO, the selection of residual covariance values is mostly determined by
trial and error. Since the factor graph optimizer only weighs residual terms
relatively, the absolute value of covariances may not have physical meaning.
Although there exist auto-tuning methods for Kalman filters, auto-tuning for
factor graphs remains unexplored. Besides, due to vibrations and impacts, the
actual sensor noise model may be heteroskedastic instead of constant. A more
careful study of sensor noise characteristics in legged robot state estimation
could not only benefit legged locomotion but also shed light on other sensor
fusion problems.

\section{Conclusions}\label{tro-sec:conclusion}
In this paper, we studied a multi-sensor visual-inertial-leg odometry hardware
and algorithm solution that fuses information from one camera, multiple
inertial sensors, and multiple motor encoders. The inclusion of inertial
measurement units on robot feet significantly improves the odometer accuracy.
By comparing various formulations of visual-inertial-leg odometry in different
locomotion conditions, we understand both the theory and best implementation
practices. In addition to this paper, we open-source our code implementation
Cerberus 2.0 along with a large dataset.

\appendix[]\label{sec:appendix}
\rev{The derivation of the process Jacobian has been presented in the VIO and legged robot state-estimation literature such as \cite{wisth2022vilens, elnecavexavier:hal-04382871, yang2022cerberus}. Here, we want to emphasize that Equation \ref{tro-eqn:quat-exp-map} and Equation \ref{tro-eqn:rot-exp-map} have first-order equivalence because both can lead to the same small rotation error dynamics. Then we will use Equation \ref{tro-eqn:quat-exp-map} to derive two key terms in the measurement Jacobian. }

\subsection{\rev{Small Rotation Error Dynamics}}
\rev{We define true orientation as $\bm{q}$ and our estimation as  $\hat{\bm{q}}$, so the estimation error quaternion is $\delta \bm{q} = \hat{\bm{q}}^{-1}\otimes\bm{q}$. From standard quaternion kinematics \cite{qin2018vins}}
\begin{equation}
    \dot{\bm{q}} =  \frac{1}{2} \bm{q} \otimes \begin{bmatrix} \bm{\omega} - \bm{b}_g\\ 0 \end{bmatrix}
\end{equation}
\rev{we have}
\begin{align}
    (\hat{\bm{q}} \dot{\otimes} \delta \bm{q}) =
    \dot{\hat{\bm{q}}}\otimes\delta \bm{q} + \hat{\bm{q}}\otimes \delta \dot{\bm{q}} =
    \frac{1}{2} \hat{\bm{q}}\otimes \delta \bm{q} \otimes \begin{bmatrix} \bm{\omega} - (\hat{\bm{b}}_g+\delta \bm{b}_g)\\ 0 \end{bmatrix}
\end{align}
\rev{which can be simplified to }
\begin{align}
    2\delta \dot{\bm{q}} & =  \mathcal{R}(\begin{bmatrix} \bm{\omega} - (\hat{\bm{b}}_g+\delta \bm{b}_g))\\ 0 \end{bmatrix})\delta \bm{q} - \mathcal{L}(\begin{bmatrix} \bm{\omega} - \hat{\bm{b}}_g\\ 0 \end{bmatrix})\delta \bm{q}\nonumber \\
                         & =   \begin{bmatrix}
                                   - \skewcross{2(\bm{\omega}-\hat{\bm{b}}_g)+\delta \bm{b}_g} & -\delta \bm{b}_g \\
                                   \delta \bm{b}_g                                             & 0
                               \end{bmatrix}
    \delta \bm{q}
\end{align}
\rev{Because} $\delta \bm{q} = \text{Exp}(\delta \bm{\theta}) = \begin{bmatrix}
        \frac{1}{2}\delta \bm{\theta} \\ 1
    \end{bmatrix}$\rev{, then} $\delta \dot{\bm{q}} = \begin{bmatrix}
        \frac{1}{2}\delta \dot{\bm{\theta}} \\ 0
    \end{bmatrix}$\rev{. Ignoring any product of two error terms, we get }
\begin{align}
    \begin{bmatrix}
        \delta \dot{\bm{\theta}} \\ 0
    \end{bmatrix} & =
    \begin{bmatrix}
        - \skewcross{2(\bm{\omega}-\hat{\bm{b}}_g)+\delta \bm{b}_g} & -\delta \bm{b}_g \\
        \delta \bm{b}_g                                             & 0
    \end{bmatrix}
    \begin{bmatrix}
        \frac{1}{2}\delta \bm{\theta} \\ 1
    \end{bmatrix}                                                                                                                              \\
    \delta \dot{\bm{\theta}}                 & = - \skewcross{\bm{\omega}-\hat{\bm{b}}_g}\delta \bm{\theta} - \delta \bm{b}_g \ \label{eqn:quat-error-delta-theta}
\end{align}

\rev{Alternatively, the result can be derived using rotation matrices. Write $\text{Exp}(\delta \bm{\theta}) = I + \skewcross{\delta \bm{\theta}}$. We notice }
\begin{align}
    R = \hat{R}(I + \skewcross{\delta \bm{\theta}}) = \hat{R} + \hat{R}\skewcross{\delta \bm{\theta}}
\end{align}
\rev{then }
\begin{align}
    \dot{R} = \dot{\hat{R}} + \dot{\hat{R}}\skewcross{\delta \bm{\theta}} + \hat{R}\skewcross{\delta \dot{\bm{\theta}}}
\end{align}
\rev{Then we can expand $\dot{R} =  R \skewcross{\bm{\omega}-\bm{b}_g}$, the matrix form of the rotation dynamics ~\cite{lynch2017modern} as}
\begin{align}
    \dot{\hat{R}} + \dot{\hat{R}}\skewcross{\delta \bm{\theta}} + \hat{R}\skewcross{\delta \dot{\bm{\theta}}} & =  (\hat{R} + \hat{R}\skewcross{\delta \bm{\theta}})\skewcross{\bm{\omega}-\bm{b}_g}                   \\
    \skewcross{\delta \dot{\bm{\theta}}}                                                                      & = \skewcross{\delta \bm{\theta}}\skewcross{\bm{\omega}-\hat{\bm{b}}_g} \nonumber                       \\
                                                                                                              & - \skewcross{\bm{\omega}-\hat{\bm{b}}_g} \skewcross{\delta \bm{\theta}}  - \skewcross{\delta \bm{b}_g}
\end{align}
\rev{Since $\skewcross{\bm{a}\times \bm{b}} =\skewcross{\bm{a}}\skewcross{\bm{b}}-\skewcross{\bm{b}}\skewcross{\bm{a}}$, then again }
\begin{align}
    \skewcross{\delta \dot{\bm{\theta}}} & = \skewcross{-(\bm{\omega}-\hat{\bm{b}}_g)\times \delta \bm{\theta}} - \skewcross{\delta \bm{b}_g} \\
    \delta \dot{\bm{\theta}}             & = -\skewcross{\bm{\omega}-\hat{\bm{b}}_g} \delta \bm{\theta} - \delta \bm{b}_g,
\end{align}
\rev{The result is the same as Equation \ref{eqn:quat-error-delta-theta}.}
\subsection{\rev{Measurement Jacobian Terms}}
\rev{For term in Equation \ref{tro-eqn:mipo-measurement}}
\begin{equation*}
    \bm{h}_2 =     \text{Log}(\bm{q}(\bm{\alpha})^{-1}\otimes \hat{\bm{q}}^{-1}\otimes \hat{\bm{q}}_f)
\end{equation*}
\rev{First, we calculate the body orientation derivative. }
\begin{align}
    \frac{\partial \bm{h}_2}{\partial \delta \theta}= & \lim_{\delta \theta \to 0} \frac{1}{\delta
        \theta}(\text{Log}(\bm{q}(\bm{\alpha})^{-1}\otimes
    [\hat{q}\otimes\text{Exp}(\delta \theta)]^{-1}\otimes \hat{q}_f) \nonumber                     \\  &
       - \text{Log}(\bm{q}(\bm{\alpha})^{-1}\otimes \hat{q}^{-1}\otimes \hat{q}_f))
    \nonumber                                                                                      \\ = & \lim_{\delta \theta \to 0} \frac{1}{\delta \theta}
       (\text{Log}(\bm{q}(\bm{\alpha})^{-1}\otimes \text{Exp}(\delta
       \theta)^{-1}\otimes \hat{q}^{-1}\otimes \hat{q}_f) \nonumber
\end{align}
\begin{align}
      & - \text{Log}(\bm{q}(\bm{\alpha})^{-1}\otimes \hat{q}^{-1}\otimes \hat{q}_f)) \nonumber \\
    = & \lim_{\delta \theta \to 0}\frac{1}{\delta \theta}
    (2\bigg[
    \mathcal{R}(\hat{q}^{-1}\otimes \hat{q}_f))
    \mathcal{L}(\bm{q}(\bm{\alpha})^{-1})\begin{bmatrix}
                                             -\frac{1}{2}\delta \theta \\
                                             1
                                         \end{bmatrix} \nonumber                             \\
      & -
    \mathcal{R}(\hat{q}^{-1}\otimes \hat{q}_f)
    \mathcal{L}(\bm{q}(\bm{\alpha})^{-1})\begin{bmatrix}
                                             0 \\
                                             1
                                         \end{bmatrix}
    \bigg]_{3\times3}) \nonumber
    \\
    = & \lim_{\delta \theta \to 0}\frac{
    -[\mathcal{R}(\hat{q}^{-1}\otimes \hat{q}_f)
    \mathcal{L}(\bm{q}(\bm{\alpha})^{-1})]_{3\times3}\delta \theta
    }{\delta \theta}.
\end{align}
\rev{An interesting property of the multiplicative maps is }
$
    \mathcal{L}(q)_{3\times3} = \mathcal{R}(q^{-1})_{3\times3}.
$ So
\begin{align}
    \frac{\partial \bm{h}_2}{\partial \delta \theta} & = -[\mathcal{R}(\hat{q}^{-1}\otimes \hat{q}_f)
    \mathcal{L}(\bm{q}(\bm{\alpha})^{-1})]_{3\times3}                                                 \\  & =
       -[\mathcal{L}(\hat{q}_f^{-1}\otimes \hat{q})
    \mathcal{R}(\bm{q}(\bm{\alpha}))]_{3\times3}.
\end{align}
\rev{Secondly, we compute the foot orientation derivative. }
\begin{align}
    \frac{\partial \bm{h}_2}{\partial \delta \theta_f} = & \lim_{\delta \theta_f \to 0} \frac{1}{\delta
        \theta_f}(\text{Log}(\bm{q}(\bm{\alpha})^{-1}\otimes \hat{q}^{-1}\otimes
    \hat{q}_f\otimes\text{Exp}(\delta \theta_f)) \nonumber                                                                                         \\  & -
    \text{Log}(\bm{q}(\bm{\alpha})^{-1}\otimes \hat{q}^{-1}\otimes \hat{q}_f))                                                                     \\ =
                                                         & \lim_{\delta \theta_f \to 0} \frac{1}{\delta
        \theta_f}(2\bigg[\mathcal{L}(\bm{q}(\bm{\alpha})^{-1}\otimes
        \hat{q}^{-1}\otimes \hat{q}_f)
    \begin{bmatrix}
            \frac{1}{2}\delta \theta_f \\
            1
        \end{bmatrix} \nonumber                                                                                                                     \\
                                                         & - \mathcal{L}(\bm{q}(\bm{\alpha})^{-1}\otimes \hat{q}^{-1}\otimes \hat{q}_f)
        \begin{bmatrix}
            0 \\
            1
        \end{bmatrix}\bigg]_{3\times 3})
    \\
    =                                                    & \mathcal{L}(\bm{q}(\bm{\alpha})^{-1}\otimes \hat{q}^{-1}\otimes \hat{q}_f)_{3\times 3}.
\end{align}

\bibliographystyle{IEEEtran}
\bibliography{reference}

\end{document}